\documentclass[runningheads]{llncs}

\usepackage{eccv}
\usepackage{eccvabbrv}

\usepackage{graphicx}        % 图片处理
\usepackage{booktabs}        % 表格美化（仅加载1次）
\usepackage[accsupp]{axessibility}  % PDF可访问性
\usepackage{orcidlink}       % ORCID链接

\usepackage{amsmath,amssymb}

\usepackage{pifont}          % 特殊符号（仅加载1次）
\usepackage{multirow}        % 跨行表格
\usepackage{makecell}        % 单元格换行
\usepackage{array}           % 增强表格功能
\usepackage[table]{xcolor}   % 表格颜色（包含colortbl，无需单独加载）

\usepackage{enumitem}

\usepackage{algorithm}
\usepackage{algorithmic}

\usepackage{textcomp}
\usepackage{subcaption}

\usepackage{threeparttable} 

\usepackage[utf8]{inputenc}
\usepackage[T1]{fontenc}
\usepackage[most]{tcolorbox}

\usepackage{tabularx}       % 自动分配列宽
\usepackage{ragged2e}       % 解决文本对齐问题
\usepackage{listings}

\usepackage{hyperref}

\title{WebRetriever: A Large-Scale Comprehensive Benchmark for Efficient Web Agent Evaluation} 
\titlerunning{WebRetriever}

\author{
Wei Dong\inst{1}$^{*}$\orcidlink{https://orcid.org/0009-0001-1085-8735} \and
Tianyu Fu\inst{1}$^{*}$\orcidlink{https://orcid.org/0009-0001-7656-350X} \and
Zhe Yu\inst{1}\orcidlink{https://orcid.org/0009-0001-7896-8412} \and
Hanning Wang\inst{1}\orcidlink{https://orcid.org/0009-0003-1474-3657} \and
Anyang Su\inst{1}\orcidlink{https://orcid.org/0000-0002-0085-3527} \and
\\
Zhizhou Fang\inst{1}\orcidlink{https://orcid.org/0009-0002-7141-7121} \and
Yuyang Chen\inst{1}\orcidlink{https://orcid.org/0009-0006-0256-4979} \and
Shuo Wang\inst{1}\orcidlink{https://orcid.org/0009-0003-9907-6186} \and
Minghui Wu\inst{1}\orcidlink{https://orcid.org/0009-0001-4577-3002} \and
\\
Ping Jiang\inst{1}\orcidlink{https://orcid.org/0009-0008-6161-060X} \and
Zhen Lei\inst{2,3}\orcidlink{https://orcid.org/0000-0002-0791-189X} \and
Chenxu Zhao\inst{1}$^{\dagger}$\orcidlink{https://orcid.org/0000-0003-4044-5701}
}
\authorrunning{W.~Dong, T.~Fu et al.}

\institute{
Mininglamp Technology\\
\and
School of Artificial Intelligence, University of Chinese Academy of Sciences
\and
MAIS, Institute of Automation, Chinese Academy of Sciences
\\
\email{zhaochenxu@mininglamp.com}
\\
\url{https://github.com/Mininglamp-AI/WebRetriever}
}

\begin{document}
\maketitle
\renewcommand{\thefootnote}{}
\footnotetext{$^*$Equal contribution. $^\dagger$Corresponding author.}
\renewcommand{\thefootnote}{\arabic{footnote}}

\begin{abstract}
  As web agents increasingly demonstrate capabilities in automated task execution, the development of robust evaluation frameworks for assessing their navigation and task completion performance has emerged as a critical research priority. However, existing benchmarks exhibit several fundamental limitations. First, they suffer from insufficient scale and limited domain diversity, thereby constraining comprehensive evaluation of cross-domain generalization. Second, prevailing LLM-as-Judge evaluation methodologies inadequately capture fine-grained interaction semantics, particularly regarding precise query formulation and filtering operations. Third, current benchmarks predominantly emphasize navigation success metrics while neglecting critical requirements for real-world deployment scenarios. To address these limitations, we introduce WebRetriever, a large-scale benchmark encompassing 800 websites and 1,550 tasks across diverse domains, including consumer, professional, and enterprise sectors, with comprehensive coverage of user intent patterns. We propose NavEval (Navigation Evaluation), a novel LLM-as-Judge framework that leverages rich interaction context beyond visual screenshots, achieving state-of-the-art alignment with human judgment across multiple evaluation datasets. Furthermore, we establish three complementary evaluation protocols that collectively provide holistic assessment of web agent capabilities: navigation proficiency, knowledge-assisted interaction, and end-to-end task completion with information extraction. Extensive experimental analysis reveals substantial performance disparities across evaluation protocols, demonstrating that navigation success alone serves as an insufficient predictor of real-world application effectiveness. WebRetriever delivers fine-grained diagnostic insights into agent capabilities and establishes a rigorous foundation for advancing web agent research and development. 
\end{abstract}

\section{Introduction}
\label{section:intro}
Recent advances in large language models have significantly improved language understanding, reasoning, and decision making, driving multimodal web agents to emerge as a key paradigm for automating complex online tasks \cite{bai2025qwen2,qin2025ui,fu2025mano,lai2024autowebglm,hong2024cogagent,guo2025seed1}. By jointly perceiving the visual content, structural information, and textual semantics of web pages, these agents can interact with real websites and have demonstrated strong potential in e-commerce \cite{yao2022webshop,chen2024chatshop}, customer service \cite{deng2024multi}, and enterprise automation \cite{drouin2024workarena,boisvert2024workarena++}. As model capability and system complexity continue to grow, accurate, reliable, and scalable evaluation of web agents has become a critical bottleneck for further progress.
\par
Web agent evaluation benchmarks are divided into offline and online settings. Offline benchmarks \cite{deng2023mind2web,liu2023agentbench,kapoor2024omniact,liu2024visualwebbench,liu2024visualagentbench,lù2024weblinx,yao2022webshop,zhou2024webarena,koh2024visualwebarena,chen2024webvln,garg2025real} provide controlled, reproducible environments but suffer from limited fidelity to real-world complexity, creating gaps between benchmark scores and actual performance. Recent online benchmarks \cite{drouin2024workarena,boisvert2024workarena++,yoran2024assistantbench,fan2024videoagent,xue2025an,tian2025mmina} enable live website interaction for better real-world characterization. However, they remain limited in website scale, domain coverage, and intent diversity, failing to systematically capture the breadth required for practical deployment, leading to biased assessments and overly optimistic performance conclusions.
\begin{figure*}[!b]
    \centering
    \includegraphics[width=\textwidth]{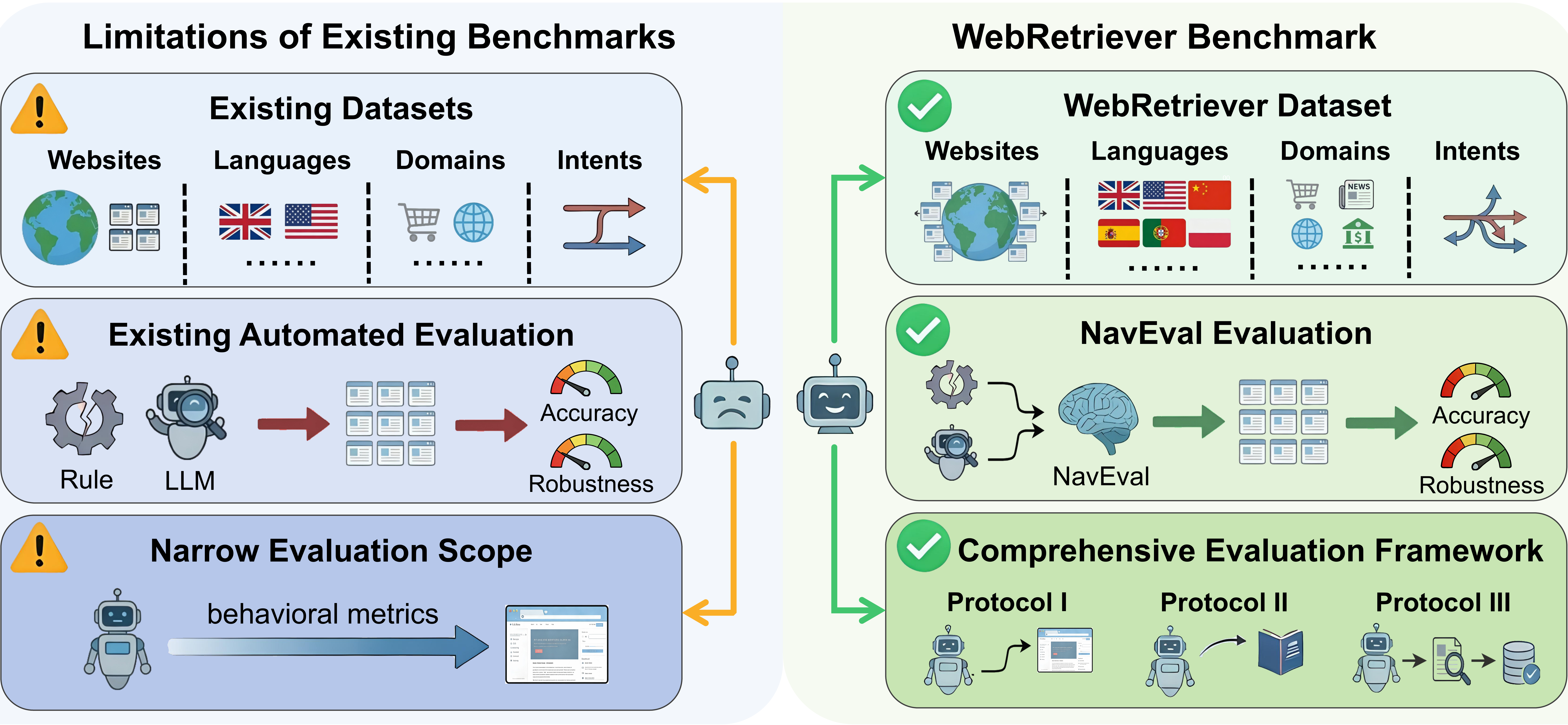}
    \caption{Motivation for the WebRetriever benchmark. WebRetriever addresses key limitations of prior work from three aspects: dataset scale and diversity, automated evaluation reliability, and deployment-oriented evaluation protocols.}
    \label{figure:motivation}
\end{figure*}
Beyond benchmark limitations, evaluation scalability presents an equally critical challenge. Quality assessment relies heavily on costly, unscalable human annotation \cite{shi2017world,zhou2024webarena,koh2024visualwebarena,pan2024webcanvas}. While automated methods \cite{he2024webvoyager,pan2024autonomous,xu2024agenttrek,xue2025an,song2025bearcubs} have been explored, they often exhibit insufficient accuracy on complex query conditions. Moreover, existing evaluation protocols assess limited dimensions. Real-world agents must navigate, leverage external knowledge, and execute end-to-end tasks like extracting information from complex websites. Current benchmarks overlook these requirements, failing to assess agents' ability to utilize external knowledge and complete full task workflows. These limitations underscore the urgent need for comprehensive benchmarks, reliable automated evaluation, and deployment-oriented protocols.
\par
To address these challenges, as illustrated in \cref{figure:motivation}, we introduce WebRetriever, a large-scale benchmark for evaluating web agents in realistic online environments. It comprises 1,550 tasks on 800 websites, covering diverse domains and user intents, and provides substantially broader website scale, domain coverage, and intent diversity than existing benchmarks. By unifying evaluation across a wide spectrum of user-intent tasks, WebRetriever mitigates biases in prior benchmarks and enables a more representative and systematic assessment of web agents’ overall capabilities. 
% To further enhance scalability and assessment accuracy, we propose NavEval, a novel automated evaluation method that substantially reduces human effort while maintaining high fidelity, achieving approximately 90\% agreement with human judgments in large-scale experiments. 
To enhance the scalability and accuracy of automated evaluation, we propose NavEval, an LLM-as-Judge method that integrates multi-source agent interaction information, achieving over 90\% agreement with human judgments in large-scale experiments.
We design three complementary evaluation protocols for comprehensive assessment: (1) Protocol I evaluates basic navigation ability to reach target pages; (2) Protocol II assesses navigation performance when provided with operational knowledge; (3) Protocol III measures end-to-end task completion by jointly evaluating navigation and information extraction, avoiding the limitation of equating page arrival with task success.
Extensive experiments validate WebRetriever, NavEval, and our evaluation protocols, providing fine-grained capability insights and establishing a foundation for advancing web agent development.
\par
In summary, the main contributions of this work are as follows:
\begin{enumerate}[noitemsep, topsep=0pt, partopsep=0pt, parsep=0pt]
\item \textbf{A large-scale, comprehensive benchmark for realistic web agent evaluation:}
We curate 1,550 tasks across 800 real websites spanning diverse domains and user intents. Compared with prior benchmarks, WebRetriever provides unprecedented scale, diversity, and coverage, enabling more comprehensive and representative evaluation of web agents in realistic online environments.
\item \textbf{A general and high-precision automated evaluation method:}
We propose NavEval, an automated evaluation method that attains approximately 90\% human-level agreement in large-scale experiments, thereby enabling practical and reliable assessment of web agent performance at scale and in real-time.
\item \textbf{Comprehensive evaluation framework:}
We propose three complementary evaluation protocols to systematically assess web agents, explicitly disentangling navigation success from answer correctness and characterizing behavioral reliability under injected operational knowledge, thereby providing diagnostic signals missing from prior benchmarks.
\end{enumerate}

\section{Related Work}
\label{section:related}
\subsection{Benchmarks for Web Agents}
\label{section:related:benchmark}
Web agent evaluation benchmarks are commonly divided into offline and online settings. Offline environments provide controlled and reproducible setups for stable capability assessment. 
Early environments such as MiniWoB~\cite{shi2017world}, MiniWoB++~\cite{liu2018reinforcement}, and CompWoB~\cite{furuta2023language} provide controllable synthetic interactions, but the substantial domain gap from real websites limits their ability to reflect real-world generalization. Subsequent efforts, including WebShop~\cite{yao2022webshop}, WebArena~\cite{zhou2024webarena}, VisualWebArena~\cite{koh2024visualwebarena}, ST-WebAgentBench~\cite{levy2024st} and Wonderbread~\cite{wornow2024wonderbread}, construct synthetic environments from real website snapshots, partially alleviating this issue, but their limited website coverage restricts robustness evaluation. Mind2Web~\cite{deng2023mind2web} and WebLINX~\cite{lù2024weblinx} adopt offline evaluation based on real HTML snapshots, enabling faster iteration and broader coverage; however, offline settings struggle to capture agents' exploration behavior and robustness to dynamic changes. Consequently, recent work has increasingly shifted toward online benchmarks. More recent benchmarks—including WorkArena~\cite{drouin2024workarena}, WorkArena++~\cite{boisvert2024workarena++}, Mind2Web-Live~\cite{pan2024webcanvas}, AssistantBench~\cite{yoran2024assistantbench}, WebVoyager~\cite{he2024webvoyager}, Bearcubs~\cite{song2025bearcubs}, and Online-Mind2Web~\cite{xue2025an}—support live evaluation and further mitigate the realism gap. However, existing online benchmarks remain limited in website scale, domain diversity, and coverage of user intent. 
% As a result, the performance reported by current benchmarks may overestimate agents’ real-world capabilities, leaving a noticeable gap between benchmark scores and practical deployment performance.

\subsection{Automatic Evaluation for Web Agents}
\label{section:related:evaluation_method}
Offline environments provide stable, reproducible settings, while online evaluation faces challenges from dynamic websites \cite{xue2025an}. SeeAct \cite{zheng2024gpt} pioneered human evaluation on live sites, but manual assessment does not scale with increasing task complexity. This motivates automated evaluation methods, which fall into rule-based \cite{zhou2024webarena,pan2024webcanvas} and LLM-as-a-judge methods \cite{zheng2023judging,li2023alpacaeval,fernandes2023devil,bai2023benchmarking}. Rule-based methods, such as Mind2Web-Live \cite{pan2024webcanvas} and AssistantBench \cite{yoran2024assistantbench}, suffer from brittleness to webpage dynamics and demand continuous maintenance, whereas LLM-as-judge approaches constrained to single-modality inputs demonstrate limited evaluation accuracy. Early approaches, such as Pan et al. \cite{pan2024autonomous}, focus on final screenshot, providing only coarse-grained assessments and missing intermediate interactions. WebVoyager \cite{he2024webvoyager} extends evaluation to full trajectories, though at substantial token cost. AgentTrek \cite{xu2024agenttrek} incorporates task descriptions and action traces to provide richer context, yet hallucinations still limit agreement with human judgments. WebJudge \cite{xue2025an} generates key steps and extracts key screenshots for LLM judgment, but the quality of key step generation and matching precision directly constrain the final accuracy.

\subsection{Evaluation Protocols for Web Agents}
\label{section:related:evaluation_protocol}
Current benchmarks predominantly assess web navigation success rates \cite{koh2024visualwebarena,drouin2024workarena,boisvert2024workarena++,yoran2024assistantbench,pan2024webcanvas,liu2024visualagentbench,xue2025an}, step correctness \cite{deng2023mind2web,kapoor2024omniact}, answer accuracy \cite{song2025bearcubs}, and execution efficiency \cite{kara2025waber} under settings where users provide only task objectives. However, in real-world scenarios, agents must leverage external knowledge and fulfill end-to-end user queries, encompassing both navigation and information extraction. Existing benchmarks fail to assess agents' navigation capabilities when augmented with external knowledge, nor can they evaluate whether end-to-end execution retrieves correct and complete data.

\section{WebRetriever}
\label{section:webretriever}
To address the limitations discussed in \cref{section:related}, this chapter presents WebRetriever's construction methodology and evaluation framework. We first describe our large-scale benchmark dataset composition. Next, we detail NavEval, our automated evaluation framework that integrates multi-modal interaction signals, applies rule-based filtering, and leverages LLMs for final assessment. Finally, we introduce three complementary evaluation protocols that emulate real-world deployment scenarios while providing multi-dimensional capability assessment.
\par

\newcommand{\yesmark}[1]{\textcolor{green!90!black}{$\scalebox{1.4}{\checkmark}$}}
\newcommand{\nomark}[1]{\textcolor{red!90!black}{\scalebox{1.4}{\ding{55}}}}
\begin{table}[!t]
  \centering
  \caption{Comparison between WebRetriever and related benchmarks. \textbf{Intent-Type}: task intent type (\textbf{\textcolor{blue}{G}}: general, \textbf{\textcolor{orange}{P}}: professional, \textbf{\textcolor{blue}{G}\&\textcolor{orange}{P}}: both); \textbf{Setting}: the evaluation environment configuration; \textbf{Online}: whether online live connection evaluation is supported in real-world environments; \textbf{Interactive}: whether the environment allows interaction; \textbf{Websites}: number of websites; \textbf{Eval-Tasks}: number of evaluation tasks.
  Statistics are reported for the web-related evaluation subsets only.
  }
  \label{table:table1}
  \small
  \resizebox{\textwidth}{!}{
  \begin{tabular}{lcccccc}
    \toprule
    \textbf{Benchmarks} & \textbf{\makecell{\#Intent-Type}} & \textbf{\#Setting} & \textbf{\#Online} & \textbf{\#Interactive} & \textbf{\#Websites} & \textbf{\#Eval-Tasks} \\
    
    \midrule
    % ---------- General ----------
    MiniWoB\cite{shi2017world} & \textcolor{blue}{\textbf{G}} & Synthetic & \nomark{red} & \yesmark{green} & 100 & 100 \\
    MiniWoB++\cite{liu2018reinforcement} & \textcolor{blue}{\textbf{G}} & Synthetic & \nomark{red} & \yesmark{green} & 100 & 100 \\
    WebArena\cite{zhou2024webarena} & \textcolor{blue}{\textbf{G}} & Semi-real & \nomark{red} & \yesmark{green} & 4 & 812 \\
    VisualWebArena\cite{koh2024visualwebarena} & \textcolor{blue}{\textbf{G}} & Semi-real & \nomark{red} & \yesmark{green} & 3 & 910 \\
    REAL\cite{garg2025real} & \textcolor{blue}{\textbf{G}} & Semi-real & \nomark{red} & \yesmark{yes} & 11 & 112 \\
    WebLINX\cite{lù2024weblinx} & \textcolor{blue}{\textbf{G}} & Real & \nomark{red} & \nomark{red} & 155 & 1368 \\
    Mind2Web\cite{deng2023mind2web} & \textcolor{blue}{\textbf{G}} & Real & \nomark{red} & \nomark{red} & 137 & 1341 \\
    Mind2Web-Live\cite{pan2024webcanvas} & \textcolor{blue}{\textbf{G}} & Real & \yesmark{green} & \yesmark{green} & 46 & 104 \\
    MMInA\cite{tian2025mmina} & \textcolor{blue}{\textbf{G}} & Real & \yesmark{green} & \yesmark{green} & 14 & 1050 \\
    AssistantBench\cite{yoran2024assistantbench} & \textcolor{blue}{\textbf{G}} & Real & \yesmark{green} & \yesmark{green} & 258 & 214 \\
    WebVoyager\cite{he2024webvoyager} & \textcolor{blue}{\textbf{G}} & Real & \yesmark{green} & \yesmark{green} & 15 & 643 \\
    Bearcubs\cite{song2025bearcubs} & \textcolor{blue}{\textbf{G}} & Real & \yesmark{green} & \yesmark{green} & 108 & 111 \\
    Online-Mind2Web\cite{xue2025an} & \textcolor{blue}{\textbf{G}} & Real & \yesmark{green} & \yesmark{green} & 136 & 300 \\
    \midrule
    % ---------- Professional ----------
    WebShop\cite{yao2022webshop} & \textcolor{orange}{\textbf{P}} & Semi-real & \nomark{red} & \yesmark{green} & 1 & 500 \\
    ST-WebAgentBench\cite{levy2024st}  & \textcolor{orange}{\textbf{P}} & Semi-real & \nomark{red} & \yesmark{green} & 3 & 222 \\
    Wonderbread\cite{wornow2024wonderbread} & \textcolor{orange}{\textbf{P}} & Semi-real & \nomark{red} & \yesmark{green} & 4 & 598 \\
    WorkArena\cite{drouin2024workarena} & \textcolor{orange}{\textbf{P}} & Real & \yesmark{green} & \yesmark{green} & 5 & 33 \\
    WorkArena++\cite{boisvert2024workarena++} & \textcolor{orange}{\textbf{P}} & Real & \yesmark{green} & \yesmark{green} & 5 & 682 \\
    OmniACT\cite{kapoor2024omniact}  & \textcolor{orange}{\textbf{P}} & Real & \yesmark{green} & \yesmark{green} & 27 & 736 \\

    \midrule
    % ---------- Ours ----------
    % \rowcolor{gray!10}
    \rowcolor{blue!10}
    \textbf{WebRetriever (Ours)} & \textbf{\textcolor{blue}{G}\&\textcolor{orange}{P}} & \textbf{Real} & \textbf{\yesmark{green}} & \textbf{\yesmark{green}} & \textbf{800} & \textbf{1550} \\

    \bottomrule
  \end{tabular}
  }
\end{table}

\subsection{Dataset Construction}
\label{section:webretriever:data_construct}
\begin{figure}[!t]
    \centering
    \begin{subfigure}{0.496\linewidth}
        \centering
        \includegraphics[width=\linewidth]{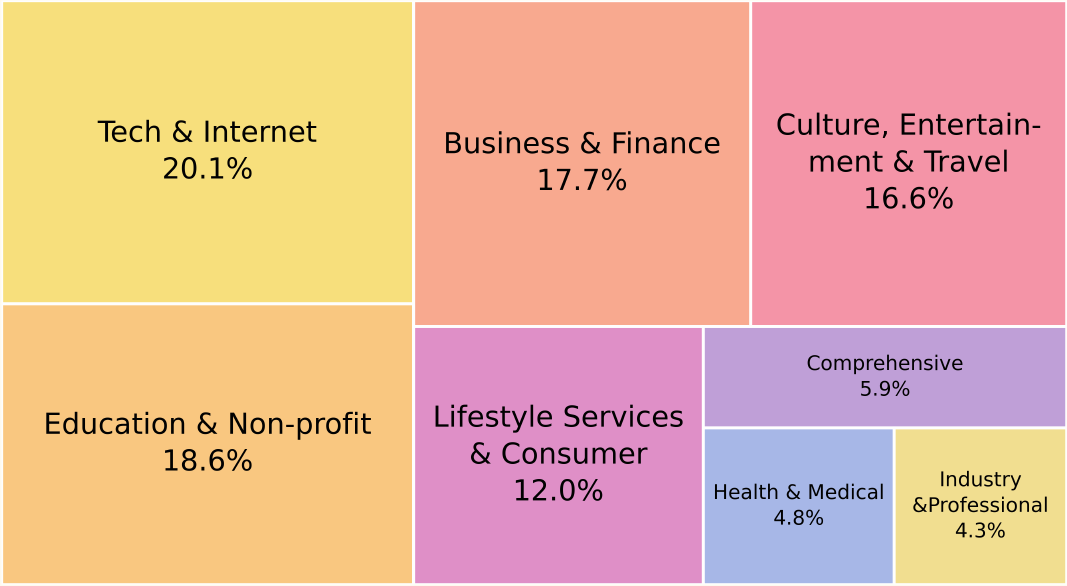}
        \caption{Domain Coverage}
    \end{subfigure}
    \hfill
    \begin{subfigure}{0.493\linewidth}
        \centering
        \includegraphics[width=\linewidth]{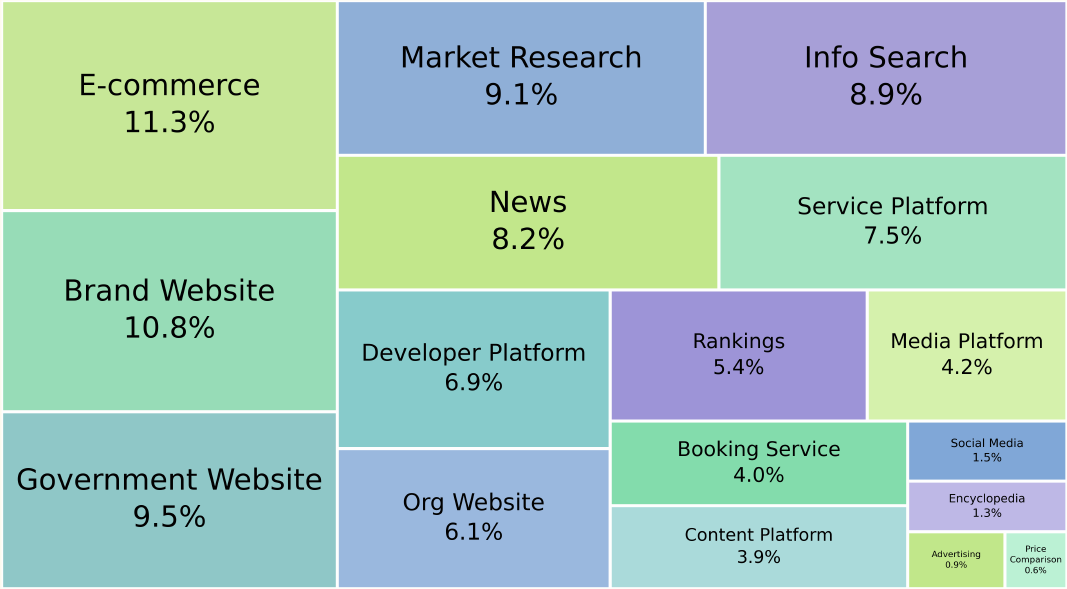}
        \caption{Website Type}
    \end{subfigure}
    \caption{Visualization of WebRetriever’s website coverage: (a) distribution across industry domains, (b) distribution of website types.}
    \label{figure:website}
\end{figure}

Given the limitations of existing benchmarks in website scale, domain coverage, and intent diversity, our WebRetriever design ensures comprehensive dataset diversity and coverage. As shown in \cref{table:table1}, compared with prior benchmarks, WebRetriever encompasses 1,550 cross-industry tasks and over 800 carefully curated high-quality active websites. To capture the fundamental landscape of mainstream internet behavior, we leverage SimilarWeb traffic data as our baseline, targeting eight core sectors including Technology \& Internet, Business \& Finance, Education \& Research, Culture, Entertainment \& Travel, Life Services, Healthcare, Industrial Manufacturing, and Public Services, as illustrated in \cref{figure:website}(a), WebRetriever spans eight core industry sectors closely related to everyday life and work.
\begin{figure}[!t]
    \centering
    \begin{subfigure}{0.496\linewidth}
        \centering
        \includegraphics[width=\linewidth]{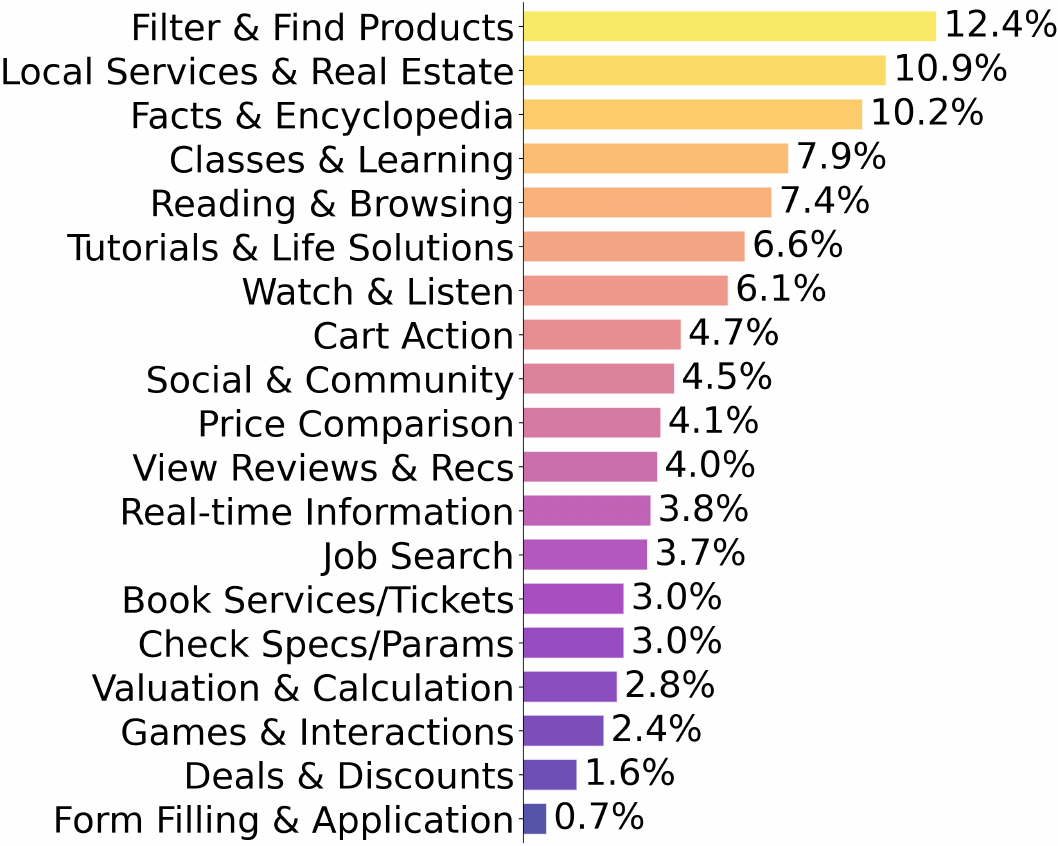}
        \caption{General User Intents}
    \end{subfigure}
    \hfill
    \begin{subfigure}{0.496\linewidth}
        \centering
        \includegraphics[width=\linewidth]{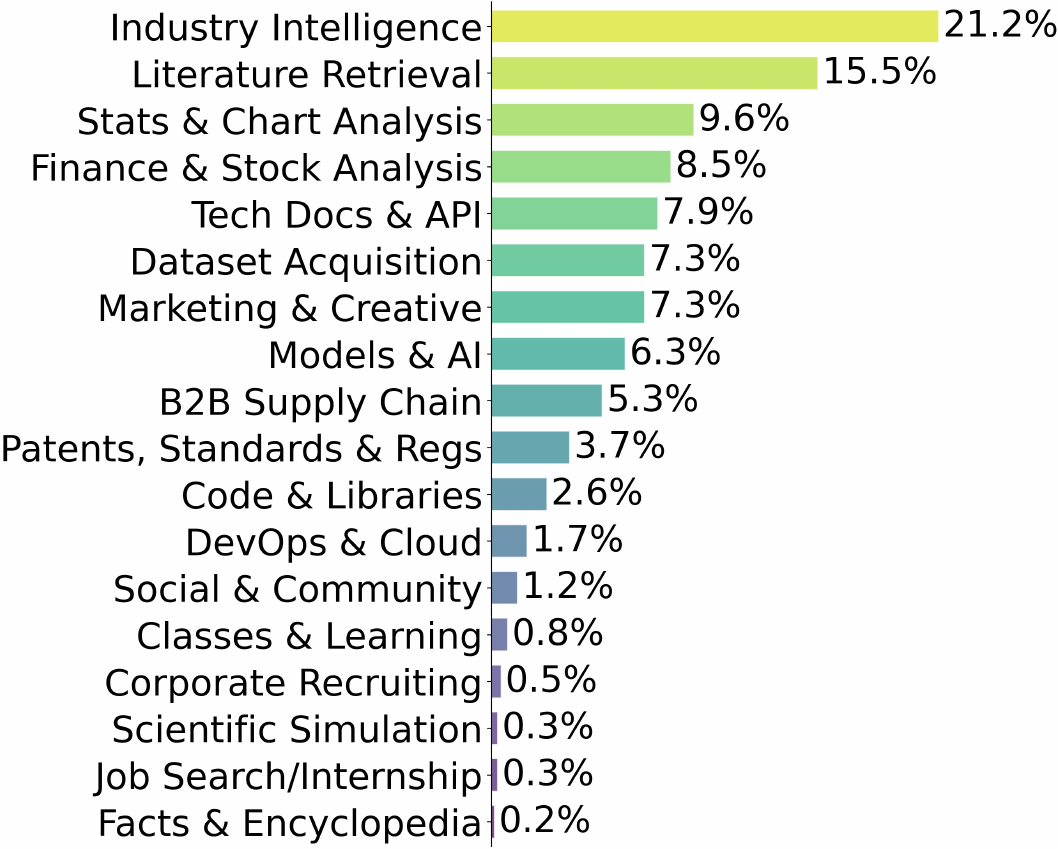}
        \caption{Professional User Intents}
    \end{subfigure}
    \caption{Visualization of user intent distributions in WebRetriever tasks: (a) distribution of general user intents, (b) distribution of professional user intents.}
    \label{figure:task}
\end{figure}
We select the top 30 sites by traffic within each sector, and additionally incorporate specialized vertical sites from authoritative internet research institutions (e.g., 199it) that maintain comprehensive data navigation repositories. This approach particularly strengthens our dataset's evaluation capabilities in business intelligence domains while ensuring broad website type distribution (\cref{figure:website}(b)).
At the task design level, a diverse team comprising industry experts, mid-level managers, senior data analysts, and university students contributes tasks for websites within their domains of expertise, grounded in authentic user intents. Building upon this foundation, we categorize task intents into two orthogonal dimensions: "general intents" and "professional intents" (\cref{figure:task}). By combining these intent types with domain-specific "Fact Elements," we generate numerous tasks with distinct scenario characteristics. This approach evaluates not only whether agents can navigate websites, but crucially, whether they understand the underlying business logic.
To ensure rigorous executability and business authenticity for each generated task, we conduct multiple rounds of cross-validation during the task refinement phase, guided by three core principles: uniqueness, stability, and logical authenticity. While traditional datasets often feature flat task structures, WebRetriever tasks must align with expert operational intuition. For instance, in the financial domain, when a task involves low-frequency trading in over-the-counter markets, expert validation led us to eliminate unrealistic requirements for "intraday high-frequency data" queries, instead directing agents toward long-term quarterly or annual reports that reflect actual financial data disclosure patterns.
Furthermore, to systematically characterize agent performance across varying difficulty levels, we categorize tasks based on the number of action steps ${Step}_{n}$ required by human annotators: tasks with $n < 6$ are defined as easy, $6 \leq n \leq 15$ as medium, and $n > 15$ as hard.
\par
Considering the dynamic nature of web content, WebRetriever will be continuously maintained: when a task becomes obsolete or unreproducible due to page updates, it will be replaced with a new task of matching difficulty, ensuring comparability and long-term validity across dataset versions.
\begin{figure*}[!b]
    \centering
    \includegraphics[width=\textwidth]{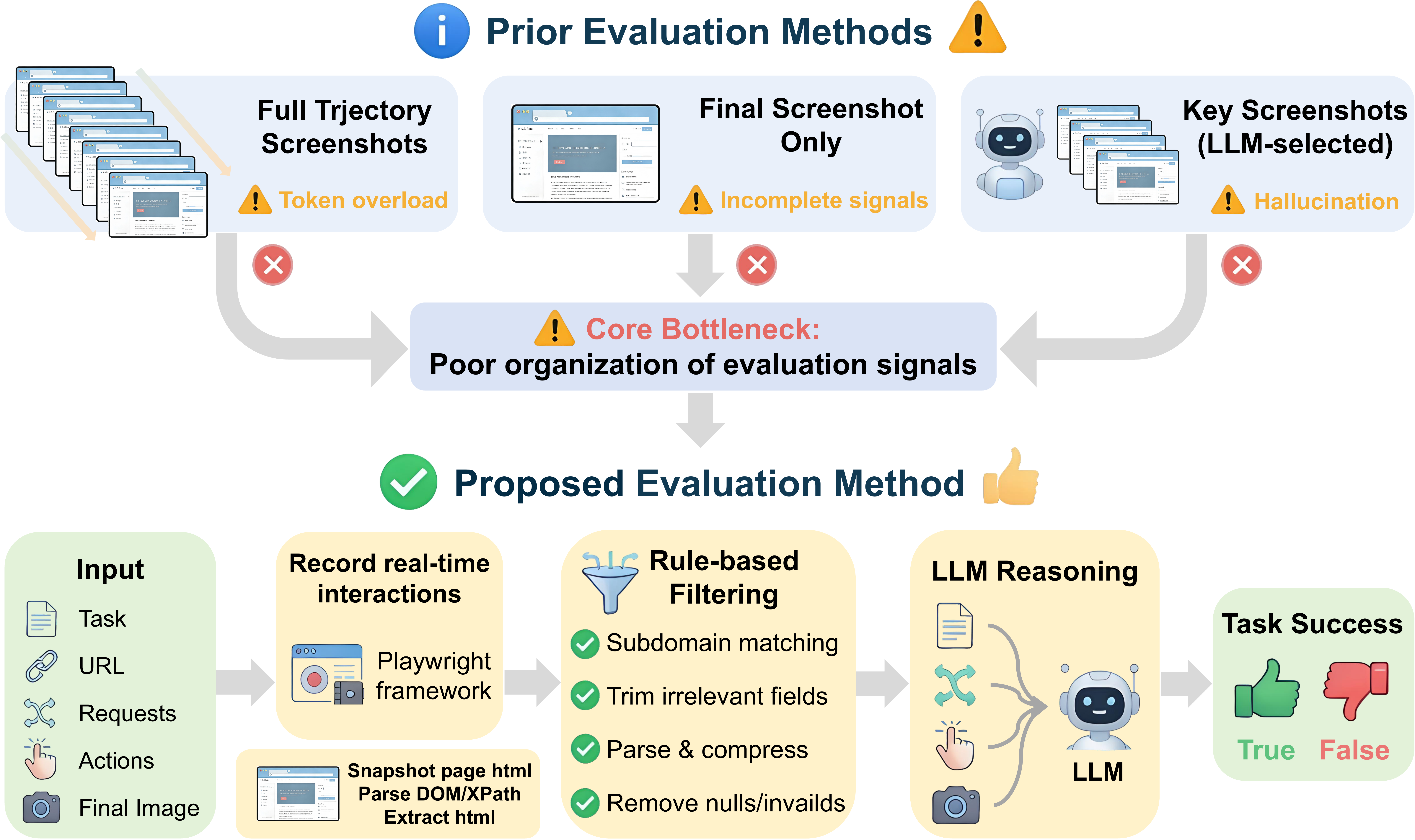}
    \caption{
    Workflow of NavEval. Compared to existing methods, NavEval applies rule-based filtering to extract fine-grained intermediate signals, which are then jointly reasoned with the task description, action trajectory, and final screenshot by an LLM to determine task success, enabling robust evaluation with higher human agreement rates.}
    \label{figure:naveval}
\end{figure*}

\subsection{NavEval}
\label{section:webretriever:naveval}
Evaluating web agents in real-world environments is crucial yet inherently challenging due to the continuous evolution of web content. Human assessment is costly and difficult to scale, while existing automated approaches—whether rule-based or LLM-based—struggle to simultaneously achieve high accuracy and robustness in dynamic online evaluation settings.
Current LLM-based evaluation pipelines exhibit limitations in organizing and utilizing evaluation signals. As illustrated in \cref{figure:naveval}, existing methods employ three screenshot strategies: (1) full-trajectory screenshots, which provide comprehensive coverage but suffer from redundancy and high computational costs; (2) final screenshot only, which captures task completion status but misses critical intermediate process information; and (3) LLM-selected key screenshots, which reduce redundancy but lack sufficient accuracy in key step selection and matching. These limitations reveal the core challenge of automated evaluation: how to extract critical information to enhance LLM judgment accuracy.
\par
To address these limitations, we propose NavEval (\cref{figure:naveval}), which integrates more diverse information from agent-browser interactions. Beyond actions and final screenshots, NavEval leverages all navigation URLs and network requests generated during interactions. Through rule-based filtering to eliminate substantial noise, these signals are restructured into organized information that enhances LLM judgment accuracy. Formally, given an input consisting of a task description $T$, a website URL $U$, a sequence of web requests $R = (r_1, r_2, ..., r_n)$, a sequence of executed actions $A = (a_1, a_2, ..., a_n)$, and the final screenshot $I$, NavEval produces a binary classification output indicating task success or failure:
\begin{equation}
    P = \text{NavEval}(T, U, R, A, I),
\end{equation}
where $P \in \{\text{True}, \text{False}\}$ denotes whether the task was successfully completed.
To robustly extract intermediate interaction information, we develop an online task evaluation framework based on Playwright, which enables real-time assessment of web agents in live environments. During task execution, the system records the web page screenshots, executed actions, and triggered requests at each step. Formally, the process can be expressed as:
\begin{equation}
    \mathcal{R}_{l}^{\prime} = \mathcal{F}_{rule}(R, U),
\end{equation}
where $\mathcal{F}_{rule}(\cdot)$ denotes the rule-based filtering function, and $\mathcal{R}_{l}^{\prime}$ represents the filtered request sequence. Specifically, NavEval first matches the triggered request set $R$ against task URL $U$ at the subdomain level to retain relevant requests. Rule-based filtering then removes task-irrelevant or random fields, normalizes structured payloads, and eliminates invalid entries, yielding a compact intermediate representation $\mathcal{R}_{l}^{\prime}$. This representation captures critical query and filtering information that images alone cannot convey. Finally, NavEval integrates the task description, action sequence, filtered intermediate signals, and final screenshot to generate a comprehensive task completion assessment.

\subsection{Evaluation Protocols}
\label{section:webretriever:protocol}
As discussed in \cref{section:related:evaluation_protocol}, existing benchmarks evaluate agents within narrow scopes, overlooking their ability to leverage external knowledge or execute critical operations after reaching target pages in realistic scenarios. Consequently, strong benchmark performance fails to translate into practical capability, revealing a significant gap between evaluation metrics and real-world utility.
\par
To comprehensively evaluate web agent capabilities, we design three protocols reflecting realistic deployment scenarios. \textbf{Protocol I} assesses fundamental navigation ability by measuring whether agents can reach target pages given user tasks. \textbf{Protocol II} evaluates navigation performance when agents are provided with operational documentation.
\begin{figure*}[!t]
    \centering
    \includegraphics[width=\textwidth]{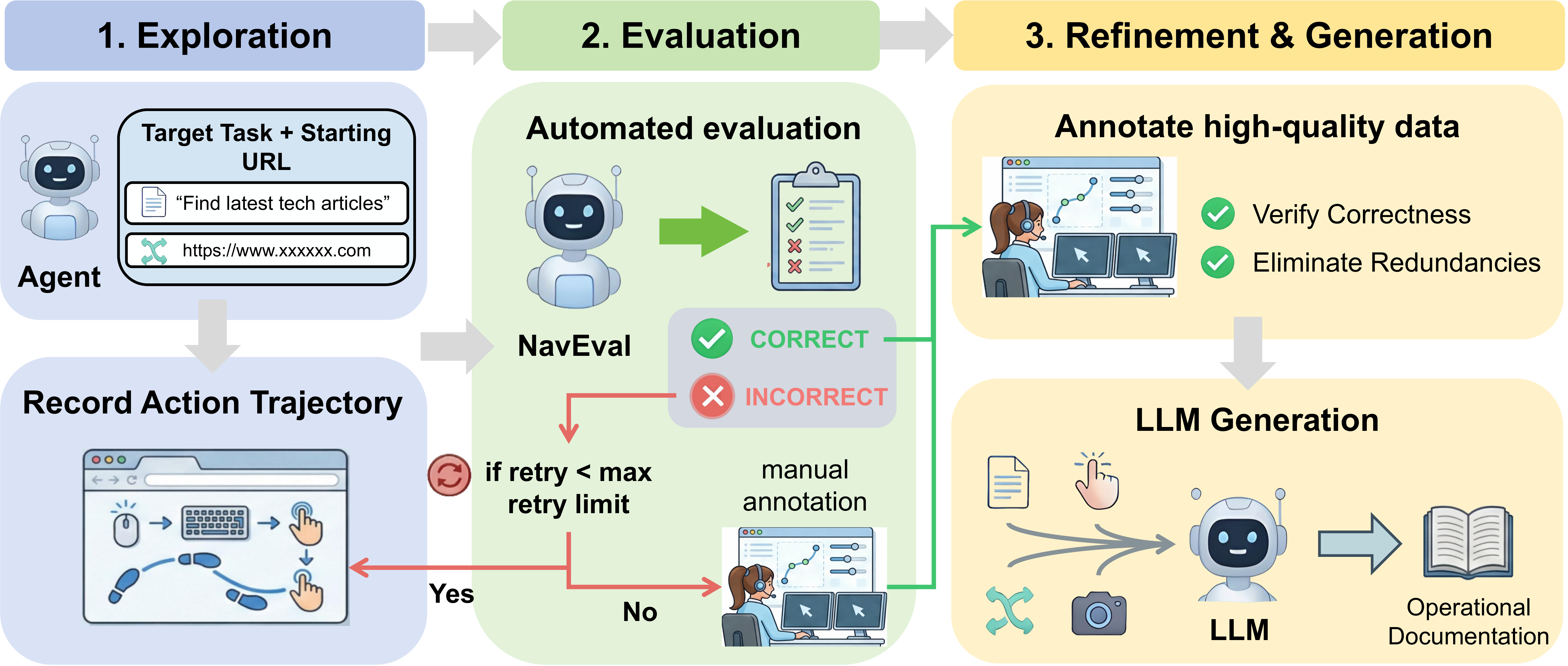}
    \caption{
    Workflow of the semi-automated pipeline for constructing operational documentation in Protocol II. The process integrates automated exploration, evaluation, manual refinement, and LLM-based generation to produce high-quality operational documentation.
    }
    \label{figure:protocol}
\end{figure*} 
For Protocol II, operational knowledge is consolidated into structured documentation through a closed-loop framework (\cref{figure:protocol}). Web agents first generate action trajectories for target tasks, which NavEval evaluates automatically. Correct trajectories proceed to annotation, while incorrect ones are regenerated until reaching retry limits, after which human annotators refine them. The annotation platform further verifies accepted trajectories and streamlines redundant operations. Finally, LLMs generate operation manuals from these refined trajectories, completing the documentation pipeline.
\par
While Protocols I and II focus on navigation, realistic web environments require capabilities beyond page navigation. Agents must demonstrate deep page understanding, multi-source information integration, and precise extraction capabilities critical for holistic evaluation. \textbf{Protocol III} mirrors realistic deployment by assessing whether agents can accurately retrieve target information across multiple modalities (text, documents, charts). Task construction follows three principles: (1) Authoritativeness: information from professional platforms ensures reliability; (2) Interaction Necessity: answers require browser-based interactions, not simple search; (3) Determinism: explicit queries yield unique, fact-based answers that remain stable for reproducible evaluation. Detailed specifications are in the supplementary material.
\par
Based on these designs, we collect 1,000, 1,000, and 100 tasks for Protocols I, II, and III, respectively. Notably, Protocols I and II share 550 overlapping tasks, with the sole difference being the availability of operational documentation.

\subsection{Evaluation Metrics}
\label{section:webretriever:metric}
To rigorously evaluate WebRetriever and NavEval, we report two metrics using human annotations as reference: Success Rate (SR) measures web agent performance across the three protocols, while Human Agreement Rate (AR) evaluates NavEval's reliability. For task set $\mathcal{T}$, with human annotation $y_t \in \{0,1\}$ and automated prediction $\hat{y}_t$:

\begin{equation}
  \text{SR} = \frac{1}{|\mathcal{T}|} \sum_{t \in \mathcal{T}} \hat{y}_t.
  \label{eq:sr}
\end{equation}

\begin{equation}
  \text{AR} = \frac{1}{|\mathcal{T}|} \sum_{t \in \mathcal{T}}
  \begin{cases}
    1, & \text{if } \hat{y}_t = y_t,\\
    0, & \text{otherwise}.
  \end{cases}
  \label{eq:ar}
\end{equation}

Higher AR indicates stronger alignment with human judgments, reflecting evaluation fidelity.

\section{Experiments and Results}
\label{section:experiment}
\subsection{Experimental Setup}
\label{section:experiment:setup}
To evaluate the effectiveness of the proposed benchmark, we develop an online task testing framework based on Playwright, enabling real-time assessment of web agents in live environments. Using this framework, we systematically evaluate a suite of state-of-the-art agents on WebRetriever across the three proposed evaluation protocols: Protocol I, Protocol II, and Protocol III. The evaluated agents include SeeAct~\cite{zheng2024gpt}, Browser-Use~\cite{browseruse}, UI-TARS-1.5~\cite{qin2025ui}, Agent-E~\cite{abuelsaad2024agent}, as well as Claude-4.5~\cite{claudesonnet45} and Gemini-2.5-Pro~\cite{comanici2025gemini} in their Computer-Use mode. To more precisely assess agent navigation and task execution in realistic web settings, we adopt a controlled experimental setup. 
% To evaluate the effectiveness of the benchmark, we develop an online task testing framework based on Playwright, enabling real-time assessment of web agents in live environments. Using this framework, we systematically evaluate a suite of state-of-the-art agents on WebRetriever across the three evaluation protocols: Protocol I, Protocol II, and Protocol III. The evaluated agents include SeeAct~\cite{zheng2024gpt}, Browser-Use~\cite{browseruse}, Agent-E~\cite{abuelsaad2024agent}, as well as Claude-4.5~\cite{claudesonnet45} and Gemini-2.5-Pro~\cite{comanici2025gemini} in their Computer-Use modes. To more precisely assess agent navigation and task execution in realistic web settings, we adopt a controlled experimental setup.
% Each task is initialized with a predefined website entry URL, while search engine access is explicitly restricted to prevent reliance on real-time retrieval shortcuts. This design ensures that the evaluation faithfully reflects the agents’ intrinsic decision-making and interaction capabilities.
Each task starts from a predefined website entry URL, with search engine access explicitly restricted to prevent reliance on real-time retrieval shortcuts. This design ensures the evaluation faithfully reflects agents’ intrinsic decision-making and interaction capabilities.
\par
Based on the execution trajectories generated by the web agents, we further assess the reliability of the automated evaluation stage by systematically comparing NavEval with representative existing methods, including Autonomous Eval~\cite{pan2024autonomous}, AgentTrek Eval~\cite{xu2024agenttrek}, WebVoyager~\cite{he2024webvoyager}, and WebJudge~\cite{xue2025an}. For a comprehensive comparison, we instantiate multiple LLM-as-a-Judge backbones, including GPT-4o~\cite{hurst2024gpt}, O4-mini~\cite{openai2025o4mini}, and Claude-4.5-Sonnet~\cite{claudesonnet45}. In our implementation, NavEval’s judgment module adopts Claude-4.5-Sonnet to ensure stable semantic understanding and reasoning. 
Further implementation details and experimental settings are provided in the supplementary material.

\subsection{Results and Analysis}
\label{section:experiment:result}
\begin{table}[!t]
  \centering
  \caption{
    Task Success Rate (SR) of web agent trajectories on WebRetriever across the three proposed evaluation protocols, assessed using NavEval and human annotation, respectively. All values are reported as percentages (\%).
  }
  \label{table:table2}
  \small
  \resizebox{0.8\textwidth}{!}{  % 调整宽度适配6列结构
  \begin{tabular}{lccccc}
    \toprule
    \multirow{2}{*}{\textbf{Agent}} & \multicolumn{2}{c}{\textbf{Protocol I}} & \multicolumn{2}{c}{\textbf{Protocol II}} & \textbf{Protocol III} \\
    \cmidrule(lr){2-3} \cmidrule(lr){4-5} \cmidrule(lr){6-6}
    & \textbf{NavEval} & \textbf{Human} & \textbf{NavEval} & \textbf{Human} & \textbf{Human} \\
    \midrule
    SeeAct & 11.3 & 9.2 & 18.9 & 17.1 & 6.0 \\
    Agent-E & 12.9 & 11.6 & 22.5 & 20.4 & 9.0 \\
    UI-TARS-1.5 & 18.3 & 16.5 & 27.0 & 24.8 & 8.0 \\
    Browser-Use & 26.6 & 24.0 & 35.2 & 31.6 & 11.0 \\
    \makecell[l]{Gemini-2.5-Pro (Computer-Use)} & 40.9 & 37.1 & 50.1 & 45.2 & 21.0 \\
    \makecell[l]{Claude-4.5 (Computer-Use)} & 31.3 & 28.1 & 40.1 & 36.3 & 16.0 \\
    \midrule
    \rowcolor{blue!10}
    \textbf{Avg SR} & \textbf{23.6} & \textbf{21.1} & \textbf{32.3} & \textbf{29.2} & \textbf{11.8} \\
    \bottomrule
  \end{tabular}
  }
\end{table}
As shown in \cref{table:table2}, human evaluation indicates that web agents perform poorly on WebRetriever across all three protocols. This sharply contrasts with their high scores on existing benchmarks and highlights the comprehensiveness and realism of WebRetriever: unlike prior benchmarks, it exposes agents to challenges closer to real-world scenarios, 
% including a large number of websites, multilingual content, multiple industry domains, and diverse user-intent tasks.
including a large number of websites spanning multiple geographic regions and industry domains, and diverse user-intent tasks.
Under Protocol I, which evaluates basic navigation skills, agents achieve an average human-assessed success rate of only 21.1\%. 
% The scale, domain diversity, multilingual coverage, and variety of user intents in WebRetriever make even seemingly straightforward path-planning tasks highly demanding. 
The scale, domain diversity, geographic breadth, and variety of user intents in WebRetriever make even seemingly straightforward path-planning tasks highly demanding. 
Compared with existing benchmarks, WebRetriever encompasses both general intent and professional intent tasks, with the latter requiring agents to navigate domain-specific websites at a level comparable to skilled human operators. This dual coverage allows for realistic evaluation of both types of capabilities within a single framework, providing critical insights into agents’ practical effectiveness.
Building on this, Protocol II introduces operational documentation to simulate agents’ ability to integrate external knowledge for automated web tasks in realistic scenarios. With access to this operational guidance, the average success rate increases to 29.2\%, with Gemini-2.5-Pro in Computer-Use mode achieving the highest score of 45.2\%. These results indicate that operational manuals can effectively support agents in navigating unfamiliar websites, reducing hallucinations and improving task completion.
Finally, Protocol III further increases the challenge by requiring end-to-end task execution, combining navigation with downstream information extraction. As shown in \cref{table:table2}, the average human-assessed success drops to 11.8\%, indicating that even when agents reach the target pages, they struggle to accurately extract and integrate the required information. Most existing benchmarks do not evaluate this capability, which is critical for realistic end-to-end deployment. Collectively, these results demonstrate that current agents remain far from reliably handling practical end-to-end web tasks, revealing a significant gap between traditional benchmark performance and real-world applicability.
\par
\begin{table}[!t]
  \centering
  \caption{
  Human Agreement Rate (AR) of web agent trajectories on WebRetriever across automated evaluation methods with different LLM-as-a-Judge models. 
  Avg AR denotes the average human agreement rate. All values are reported as percentages (\%).
  } 
  \label{table:table3}
  \resizebox{\textwidth}{!}{
  \begin{tabular}{lccccccc}
    \toprule
    \textbf{Model} & \textbf{Auto-Eval} & \textbf{SeeAct} & \textbf{Agent-E} & \textbf{Browser-Use} &
    \textbf{\makecell{Gemini-2.5-Pro\\(Computer-Use)}} & \textbf{\makecell{Claude-4.5\\(Computer-Use)}} & \textbf{Avg AR}  \\
    \midrule
    \multirow{4}{*}{GPT-4o} 
     & Autonomous Eval & 69.8 & 67.0 & 66.6 & 65.4 & 65.9 & 66.9 \\
     & AgentTrek Eval & 60.2 & 52.5 & 55.1 & 55.2 & 54.6 & 55.5 \\
     & WebVoyager & 69.1 & 66.7 & 62.2 & 64.7 & 63.1 & 65.2 \\
     & WebJudge & 71.9 & 70.5 & 72.6 & 70.8 & 70.3 & 71.2 \\
    \midrule
    \multirow{3}{*}{O4-mini} 
      & Autonomous Eval & 70.3 & 75.1 & 69.8 & 67.1 & 69.1 & 70.3 \\
      & WebVoyager & 71.6 & 75.9 & 70.5 & 67.3 & 68.6 & 70.8 \\
      & WebJudge & 76.5 & 77.6 & 75.9 & 72.5 & 73.1 & 75.1 \\
    \midrule
    \multirow{4}{*}{\makecell[l]{Claude-4.5\\-Sonnet}} 
      & Autonomous Eval & 79.5 & 76.5 & 75.2 & 74.1 & 74.4 & 75.9 \\
      & AgentTrek Eval & 66.8 & 60.7 & 61.7 & 62.4 & 61.7 & 62.7 \\
      & WebVoyager & 78.7 & 80.4 & 74.6 & 75.4 & 76.5 & 77.1 \\
      & WebJudge & 80.9 & 81.1 & 81.7 & 80.7 & 80.8 & 81.0 \\
    \midrule
    \rowcolor{blue!10}
     \textbf{\makecell[l]{Claude-4.5\\-Sonnet}}
      & \textbf{NavEval (Ours)} 
      & \textbf{92.2} & \textbf{91.3} & \textbf{90.9} & \textbf{91.4} & \textbf{90.1}
      & \textbf{91.2} \\
    \bottomrule
  \end{tabular}
  }
\end{table}
Given the substantial challenges revealed by WebRetriever, a reliable and fine-grained evaluation framework is crucial. As \cref{table:table2} shows, the task success rates evaluated by NavEval closely match those from human assessment across Protocols I and II, demonstrating both the accuracy of NavEval and its robustness across diverse task scenarios. As further illustrated in \cref{table:table3}, NavEval consistently achieves human agreement rates above 90\% across all web agents, outperforming existing methods—including Autonomous Eval, AgentTrek Eval, WebVoyager, and WebJudge—which exhibit wide variability, with average ARs ranging from 55\% to 81\%. This improvement stems not only from leveraging LLM reasoning via Claude-4.5-Sonnet, but also from incorporating fine-grained interaction data: beyond screenshots, NavEval analyzes the request sequences generated during task execution on web pages, allowing precise assessment of query execution and filtering. By combining these detailed trajectories with rule-based constraints, NavEval detects subtle behavioral differences that conventional automated methods often miss, closely approximating human judgment. These results establish NavEval as a reliable, discriminative, and fine-grained framework for evaluating web agents in realistic, instruction-guided scenarios.
\begin{table}[!h]
    \centering
    \caption{
    Human Agreement Rate (AR) of web agent trajectories on Online-Mind2Web across automated evaluation methods with different LLM-as-a-Judge models. 
    Avg AR denotes the average human agreement rate. All values are reported as percentages (\%).
    }
    \label{table:table4}
    \small
    \resizebox{0.75\textwidth}{!}{%
    \begin{tabular}{lccccc}
    \toprule
    \textbf{Model} & \textbf{Auto-Eval} & \textbf{SeeAct} & \textbf{Agent-E} & \textbf{Browser-Use} & \textbf{Avg AR} \\
    \midrule
    \multirow{4}{*}{GPT-4o} 
     & Autonomous Eval & 84.7 & 85.0 & 76.0 & 81.9 \\
     & AgentTrek Eval & 73.0 & 64.3 & 63.3 & 66.9 \\
     & WebVoyager & -- & 75.3 & 71.3 & -- \\
     & WebJudge & 86.7 & 86.0 & 81.4 & 84.7 \\
    \midrule
    \multirow{3}{*}{O4-mini} 
     & Autonomous Eval & 79.7 & 85.7 & 86.0 & 83.8 \\
     & WebVoyager & -- & 80.3 & 79.0 & -- \\
     & WebJudge & 85.3 & 86.3 & 89.3 & 87.0 \\
    \midrule
    WebJudge-7B\cite{xue2025an}
     & WebJudge & 86.0 & 87.3 & 88.3 & 87.2 \\
    \midrule
    % 修正：使用 \raisebox 调整第一列文字的垂直位置
    \raisebox{-0.55\baselineskip}{\cellcolor{blue!10}\textbf{\makecell[l]{Claude-4.5\\-Sonnet}}} 
    & \cellcolor{blue!10}\textbf{NavEval (Ours)} 
    & \cellcolor{blue!10}\textbf{96.5} 
    & \cellcolor{blue!10}\textbf{97.4} 
    & \cellcolor{blue!10}\textbf{97.1} 
    & \cellcolor{blue!10}\textbf{97.0} \\
    \bottomrule
    \end{tabular}
    }
\end{table}
To further demonstrate the generalizability and effectiveness of NavEval, we conduct additional evaluation on Online-Mind2Web~\cite{xue2025an}, a comprehensive benchmark comprising 300 high-quality real-world tasks across 136 popular websites from diverse domains. In addition to the previously used LLM-based judge models, we include WebJudge-7B, proposed in Online-Mind2Web, for a more comprehensive comparison. As shown in \cref{table:table4}, NavEval consistently outperforms existing automated evaluation methods by a clear margin on this external benchmark. Prior approaches, including Autonomous Eval, AgentTrek Eval, WebVoyager, and WebJudge, exhibit noticeable variability across evaluator backbones and agent trajectories, with average agreement rates generally below 88\%. In contrast, NavEval achieves a substantially higher Avg AR of 97\%, demonstrating strong robustness and cross-benchmark stability. This advantage stems from NavEval’s fine-grained design. By jointly leveraging rule-based constraints and structured interaction signals, particularly the request sequences generated during webpage execution, NavEval more accurately verifies query execution and filtering correctness, thereby reducing the ambiguity that commonly affects screenshot-only evaluators. Overall, the results on Online-Mind2Web confirm that NavEval generalizes effectively beyond WebRetriever and provides a reliable, high-fidelity automated evaluation framework for realistic web agent assessment.

\subsection{Ablation Analysis}
\label{section:experiment:ablation}
\begin{table}[!h]
    \centering
    \caption{
    % Ablation on operational documentation (Doc). Protocol I is originally defined without Doc, while Protocol II includes Doc. Reported values are task Success Rates (SR) of web agent trajectories on WebRetriever under different protocol settings. Settings indicate whether Doc is provided. All values are reported as percentages (\%).
    Ablation study on operational documentation (Doc). Protocol I is originally defined without Doc, while Protocol II includes Doc. Reported values are task Success Rates (SR, \%) of web agent trajectories on WebRetriever under different protocol settings. Settings indicate whether Doc is provided (w/ or w/o Doc).
    }
    \label{tab:table5}
    \small
    % \resizebox{0.65\textwidth}{!}{
    \resizebox{0.60\textwidth}{!}{
    \begin{tabular}{lccc}
    \toprule
    \textbf{Protocol} & \textbf{Setting} 
    & \textbf{\makecell{Gemini-2.5-Pro\\(Computer-Use)}} 
    & \textbf{\makecell{Claude-4.5\\(Computer-Use)}} \\
    \midrule
    \multirow{2}{*}{Protocol I}
        & -- & 40.9 & 31.3 \\
        & w/ Doc & \cellcolor{blue!10}\textbf{49.2 (+8.3)} & \cellcolor{blue!10}\textbf{39.7 (+8.4)} \\
    \midrule
    \multirow{2}{*}{Protocol II}
        & -- & 50.1 & 40.1 \\
        & w/o Doc & \cellcolor{blue!10}\textbf{41.4 (-8.7)} & \cellcolor{blue!10}\textbf{31.9 (-8.2)} \\
    \bottomrule
    \end{tabular}
    }
\end{table}
To evaluate the effect of operational documentation, we conduct an ablation study under Protocols I and II (\cref{tab:table5}). Adding documentation to Protocol I, which was originally defined without it, consistently improves performance, with success rates increasing by 8.3\% for Gemini-2.5-Pro and 8.4\% for Claude-4.5 in Computer-Use mode. Conversely, removing documentation from Protocol II, originally defined with it, leads to substantial drops of 8.7\% and 8.2\%, respectively. These results demonstrate that operational documentation provides crucial guidance on unfamiliar websites, helping agents reduce hallucinations and achieve more reliable task completion. At the same time, the moderate magnitude of improvement indicates that agents still face significant challenges in fully understanding and leveraging external knowledge, emphasizing the need for further advances in knowledge integration.
\begin{table}[!t]
    \centering
    \caption{
    Ablation on end-to-end task completion in Protocol III, reporting task Success Rates (SR, \%) of web agent trajectories on WebRetriever under different protocol settings. Settings indicate whether exact information extraction from webpages is required (w/ or w/o Extract).
    }
    \label{tab:table6}
    \small
    % \resizebox{0.65\textwidth}{!}{
    \resizebox{0.60\textwidth}{!}{
    \begin{tabular}{lccc}
    \toprule
    \textbf{Protocol} & \textbf{Setting} 
    & \textbf{\makecell{Gemini-2.5-Pro\\(Computer-Use)}} 
    & \textbf{\makecell{Claude-4.5\\(Computer-Use)}} \\
    \midrule
    \multirow{2}{*}{Protocol III}
        % & -- & 20.0 & 14.0 \\
        & -- & 21.0 & 16.0 \\
        % & w/o Extract & \cellcolor{blue!10}\textbf{40.0 (+20.0)} & \cellcolor{blue!10}\textbf{30.0 (+16.0)} \\
        & w/o Extract & \cellcolor{blue!10}\textbf{43.0 (+22.0)} & \cellcolor{blue!10}\textbf{34.0 (+18.0)} \\
    \bottomrule
    \end{tabular}
    }
\end{table}
We further conduct an ablation study on Protocol III, which requires end-to-end navigation and information extraction (see \cref{tab:table6}). Strikingly, even when agents successfully reach the target pages, their ability to extract the required information remains far from reliable: success rates for Gemini-2.5-Pro and Claude-4.5 in Computer-Use mode are only 43\% and 34\%, respectively. When evaluating full end-to-end task completion—combining navigation with information extraction—success rates drop almost by half, to 21\% and 16\%. This sharp decline reveals a critical blind spot in current agents: reaching the correct page does not guarantee successful task execution. Protocol III thus exposes the true difficulty of realistic end-to-end tasks, highlighting limitations largely overlooked by existing benchmarks and underscoring the importance of evaluating both navigation and actionable information processing capabilities.
In addition to the protocol-level ablations, we provide further analyses of NavEval, including judge backbone self-bias and rule-based filtering ablations, in the supplementary material.

\section{Conclusion}
\label{section:conclusion}
In this paper, we address the limitations of existing benchmarks for web agent evaluation, including insufficient scale, limited domain coverage, and lack of task diversity. To overcome these challenges, we introduce WebRetriever, a large-scale benchmark for realistic online evaluation, and NavEval, a scalable automated evaluation method that reduces human effort while maintaining high fidelity with human judgments. Building upon this foundation, we further propose three deployment-oriented evaluation protocols, namely Protocol I, Protocol II, and Protocol III, to systematically assess agents’ core navigation abilities, ability to leverage external knowledge, and end-to-end task execution capabilities, respectively. Extensive experiments demonstrate that our benchmark, evaluation method, and protocols provide fine-grained insights into agent performance, reveal capability gaps overlooked by conventional evaluations, and establish a solid foundation for the development of more capable and reliable web agents in practical settings.

\section*{Acknowledgements}
% Please insert your acknowledgments here.
This work was supported by Mininglamp Technology. We thank the annotation and engineering teams for their contributions to dataset construction and the design of the evaluation framework. We also thank Han Lin and Yuting Liao for their valuable support throughout this project.

\bibliographystyle{splncs04}
\bibliography{main}

\clearpage
\appendix

\section{Overview}
This supplementary material is structured into four main sections. First, we detail data construction, including the design of the three evaluation protocols, task creation procedures, and operational documentation generation. Second, we describe the experimental setup, covering model types and configurations, agents’ observation scopes during testing, and an ablation study of NavEval. Third, we present representative case studies from the WebRetriever benchmark. Finally, we provide the prompts used for NavEval evaluation and LLM-based operational documentation generation.

\section{Data Construction Details}
\label{sec:data}
In this section, we provide detailed descriptions of the construction procedures for the three protocols in WebRetriever. Specifically, we propose three evaluation protocols: (1) Protocol I evaluates the basic navigation ability required to reach target pages; (2) Protocol II assesses navigation performance when agents are provided with operational knowledge; and (3) Protocol III measures end-to-end task completion by jointly evaluating navigation and information extraction, thereby avoiding the limitation of treating page arrival as task success.

\subsection{Task Construction for Protocol I}
Protocol I and Protocol II are both designed to evaluate the navigation capability of agents under the WebRetriever benchmark. In Protocol I, agents must navigate to the target page based solely on the user task description, without additional guidance. In contrast, Protocol II additionally provides structured operational documentation describing the required interaction steps, simulating the use of knowledge bases in realistic scenarios. Although the evaluation settings differ, the task construction procedure for Protocol I and Protocol II remains the same. Task descriptions in both protocols are written by domain experts to ensure professional accuracy and realism.

\subsection{Task Construction for Protocol II}
\label{sec:data:protocol2}
\begin{figure*}[!t]
    \centering
    \includegraphics[width=\textwidth]{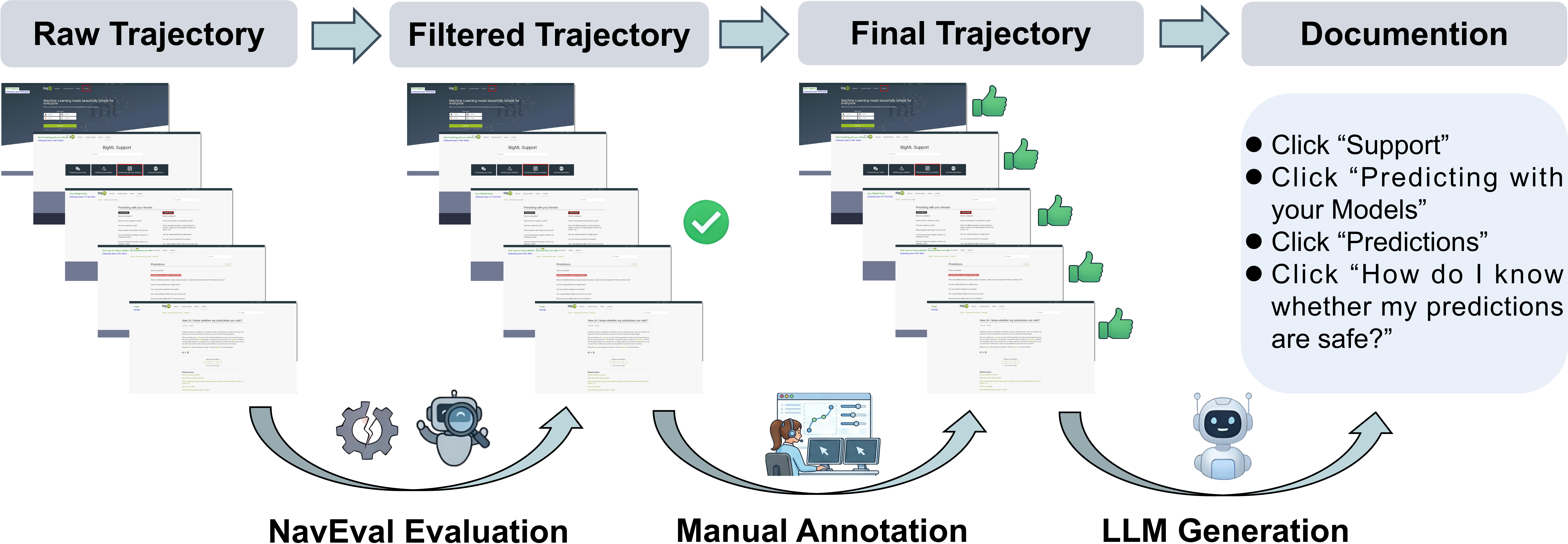}
    \caption{
    Example of the semi-automated operational documentation generation pipeline for the task "If I want to make predictions using a model, how can I ensure that my predictions are safe?" For clarity, the figure shows only the trajectory screenshots and omits other intermediate information.
    }
    \label{figure:documention}
\end{figure*}

For Protocol II, the corresponding operational documentation is generated through a semi-automated pipeline that integrates automated exploration, evaluation, manual refinement, and LLM-based generation. Within this closed-loop framework, the specified task description and the initial URL are first provided to the web agent, which explores and executes the task to generate action trajectories. These trajectories are then evaluated by NavEval using the prompts designed in \cref{sec:prompt:naveval} to determine task success. Successful trajectories are subsequently sent to a human annotation platform for refinement, producing high-quality trajectories. Trajectories that fail the NavEval check are retried, and if they still do not meet success criteria after a maximum number of attempts, they are directly refined manually to obtain high-quality trajectories. Finally, using the prompts designed in \cref{sec:prompt:documention}, Gemini-2.5-Pro generates operational documentation from these high-quality trajectories. \cref{figure:documention} illustrates this process with an example of a successful task trajectory, showing each step of the operational documentation generation pipeline.

\subsection{Task Construction for Protocol III}
Compared with the first two protocols, Protocol III evaluates whether web agents can complete user tasks end-to-end in realistic deployment scenarios, substantially increasing task difficulty. Beyond successful navigation, it also assesses whether agents can accurately integrate and extract the required information from target pages to accomplish the task. To strictly prevent agents from completing tasks using general web search engines without interacting with the specified websites, Protocol III tasks are designed as Deep Research–style end-to-end tasks. While search engines typically locate relevant sources or provide summarized results, Deep Research requires agents to enter these sources and perform systematic analysis, including reading, filtering, cross-referencing, and verifying information completeness.
\par

\setlist[itemize]{
  leftmargin=*,            % 缩进匹配示例
  label=--,                % 短横线符号，对齐示例风格
  nosep,                   % 关闭item间/parsep间距（核心）
}
\begin{table}[!t]
    \centering
    \caption{Representative Deep Research–style web interaction tasks, with example scenarios and associated challenges.}
    \label{table:table7}
    \scriptsize
    \begin{tabularx}{\linewidth}{@{} X X @{}}  % 自动分配两列宽度，消除边距冗余
    \toprule
    \textbf{Scenario Examples} & \textbf{Challenges} \\
    \midrule
    
    \rowcolor{blue!15}
    \multicolumn{2}{@{} c @{}}{\textbf{Document Extraction}} \\  % 居中+消除边距
    \begin{itemize}
        \item Financial report footnotes
        \item Engineering standards
        \item Insurance rate tables
        \item Legal judgments
    \end{itemize} & 
    \begin{itemize}
        \item Embedded content cannot be fully indexed
        \item Requires reasoning based on table headers
    \end{itemize} \\
    
    \midrule
    \rowcolor{blue!15}
    \multicolumn{2}{@{} c @{}}{\textbf{Form Interaction}} \\
    \begin{itemize}
        \item Patent searches
        \item Court docket queries
        \item Credit transfer systems
        \item Government license records
    \end{itemize} & 
    \begin{itemize}
        \item Requires simulating human operations
    \end{itemize} \\
    
    \midrule
    \rowcolor{blue!15}
    \multicolumn{2}{@{} c @{}}{\textbf{Multi-source Comparison}} \\
    \begin{itemize}
        \item Code repository diffs (PR vs Commit)
        \item Amended bills (Amendments vs Original)
        \item Environmental impact reports (Before vs After mitigation)
    \end{itemize} & 
    \begin{itemize}
        \item Must identify the latest or a specific version
        \item Must exclude outdated documents
    \end{itemize} \\
    
    \midrule
    \rowcolor{blue!15}
    \multicolumn{2}{@{} c @{}}{\textbf{Complete Data Retrieval}} \\
    \begin{itemize}
        \item Retrieve all records matching certain criteria
    \end{itemize} & 
    \begin{itemize}
        \item Answers are sets
        \item Missing any item is considered a failure
    \end{itemize} \\
    
    \midrule
    \rowcolor{blue!15}
    \multicolumn{2}{@{} c @{}}{\textbf{Multi-dimensional Chart}} \\ 
    \begin{itemize}
        \item Bidding (e.g., second-lowest bid)
        \item Sports statistics (e.g., ranking excluding certain conditions)
        \item Census data (e.g., specific demographics in a region)
        \item Multi-step filtering (e.g., among the top-5 X, identify the lowest Y)
    \end{itemize} & 
    \begin{itemize}
        \item Requires multiple analyses of raw data, reports, or charts
        \item Involves both visual reasoning and data correlation analysis
    \end{itemize} \\
    \bottomrule
    \end{tabularx}
\end{table}

As summarized in \cref{table:table7}, we categorize representative Deep Research–style tasks into several common scenarios, including document extraction, form interaction, multi-source comparison, complete data retrieval, and multi-dimensional chart analysis. These tasks require agents to navigate complex interfaces, interact with structured or semi-structured content, and reason across heterogeneous information sources. Due to the inherent challenges of these realistic end-to-end tasks, they place significant demands on web agents, requiring advanced capabilities in long-horizon planning, precise information extraction, and cross-page reasoning.
To systematically construct such challenging tasks, we deliberately introduce path obstacles during the dataset design process, forcing agents to obtain answers through genuine retrieval, reading, and reasoning rather than relying on search-engine snippets or shallow matching. This process can be organized into four stages:
\begin{itemize}[leftmargin=*, nosep]
    \item \textbf{Data Source Selection:} Appropriate data sources must be selected. Highly popular websites are generally avoided because their content is often heavily structured by search engines and easily summarized. Instead, tasks should rely on authoritative yet less exposed sources such as government portals, academic archives, professional organizations, or regulatory databases. Incorporating multilingual or region-specific websites can further increase retrieval complexity and better reflect real-world information environments.
    
    \item \textbf{Task Specification:} Tasks should target fine-grained objectives with explicit constraints. Rather than asking broad questions that can be answered directly by search engines, the task should focus on specific details embedded in documents or databases. This often involves applying multiple filtering conditions, interacting with search interfaces, or retrieving complete sets of records, which requires agents to perform multi-step navigation and information extraction.
    
    \item \textbf{Adversarial Design:} Introduce adversarial elements to increase reasoning difficulty. These may include multiple document versions, complex document layouts, or exception-based rules that require careful interpretation. Such designs ensure that successful completion depends not only on retrieval but also on accurate understanding of document structure and semantics.
    
    \item \textbf{Ground Truth Validation:} The ground truth must be carefully validated and fixed. Answers should rely on stable historical data from authoritative sources and be cross-verified to ensure correctness and uniqueness. For reliable automated evaluation, the final answers should be manually confirmed and represented in a structured form, allowing precise comparison during evaluation.
\end{itemize}

\subsection{Annotation Details} 
WebRetriever was annotated by a 14-member team comprising domain experts, managers, data analysts, and university students. The annotation process followed a five-stage pipeline consisting of task design, annotation, cross-validation, quality review, and final verification, with strict annotator--reviewer separation and 12 calibration sessions conducted throughout the annotation cycle.
Rather than relying on parallel annotation for inter-annotator agreement (IAA), we adopted a sequential multi-stage quality control process in which every task underwent three independent review stages, with mandatory revision whenever disagreements arose. The corresponding correction rates at each stage were 30\%, 26\%, and 27\%, respectively, with 54\% of tasks receiving at least one correction, demonstrating the complementary role of successive review stages.

\subsection{More Details About Dataset}
\begin{figure}[!b]
    \centering
    \begin{subfigure}{0.3\linewidth}
        \centering
        \includegraphics[width=\linewidth]{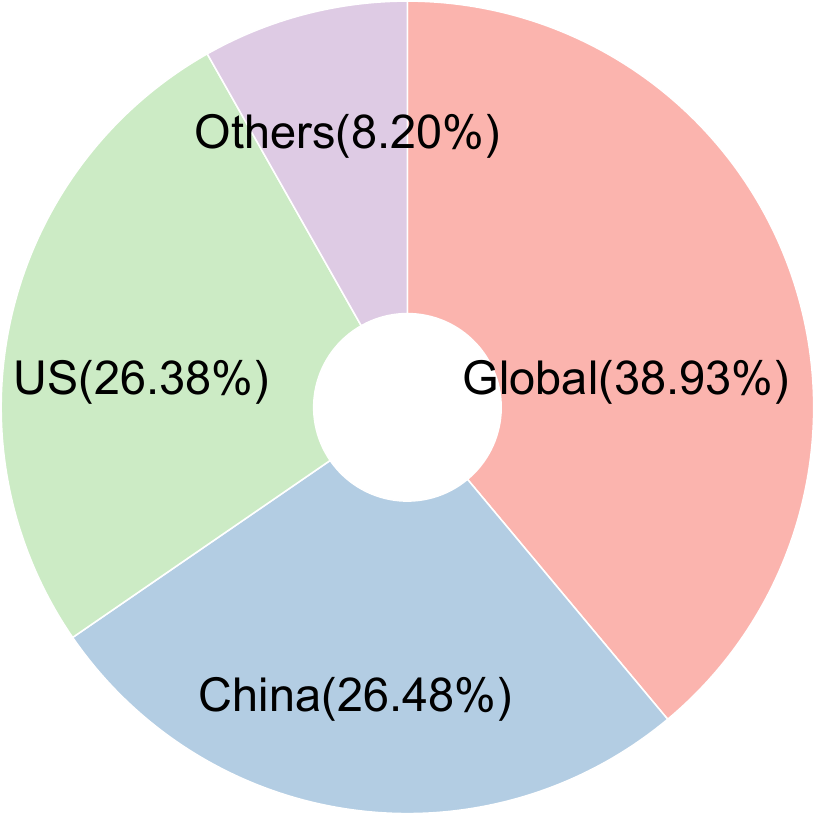}
        \caption{Geographic Distribution}
    \end{subfigure}
    % \hfill
    \hspace{0.1\linewidth} % 手动设置两张图的间距
    \begin{subfigure}{0.3\linewidth}
        \centering
        \includegraphics[width=\linewidth]{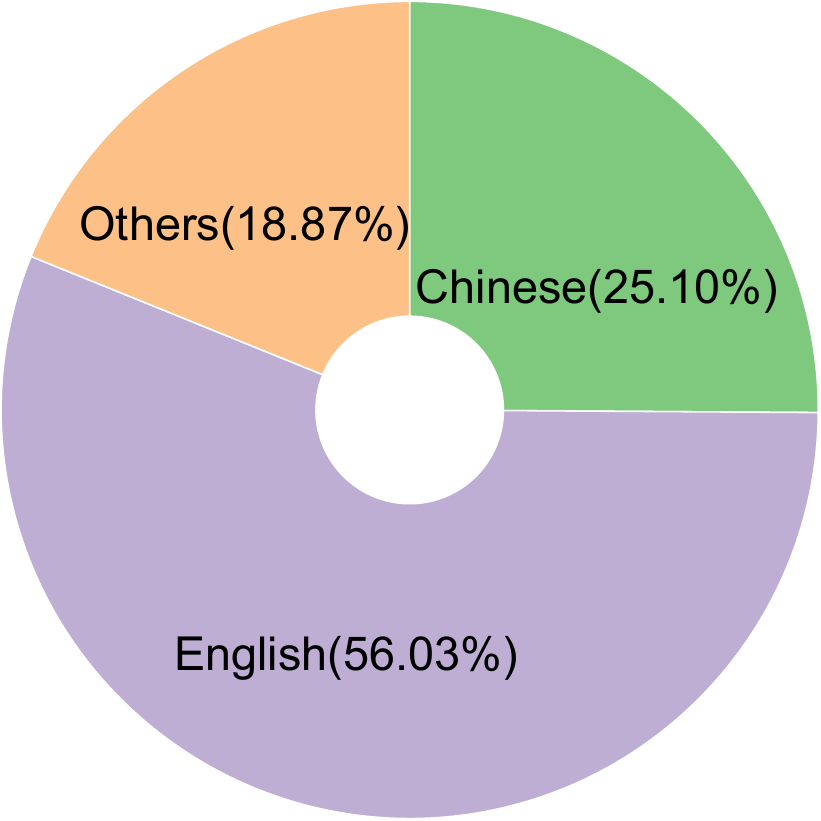}
        \caption{Language Distribution}
    \end{subfigure}
    \caption{Details of websites in the WebRetriever benchmark. (a) Geographic distribution of websites. (b) Language distribution of websites.}
    \label{figure:info}
\end{figure}
\begin{figure*}[!t]
    \centering
    \includegraphics[width=0.7\textwidth]{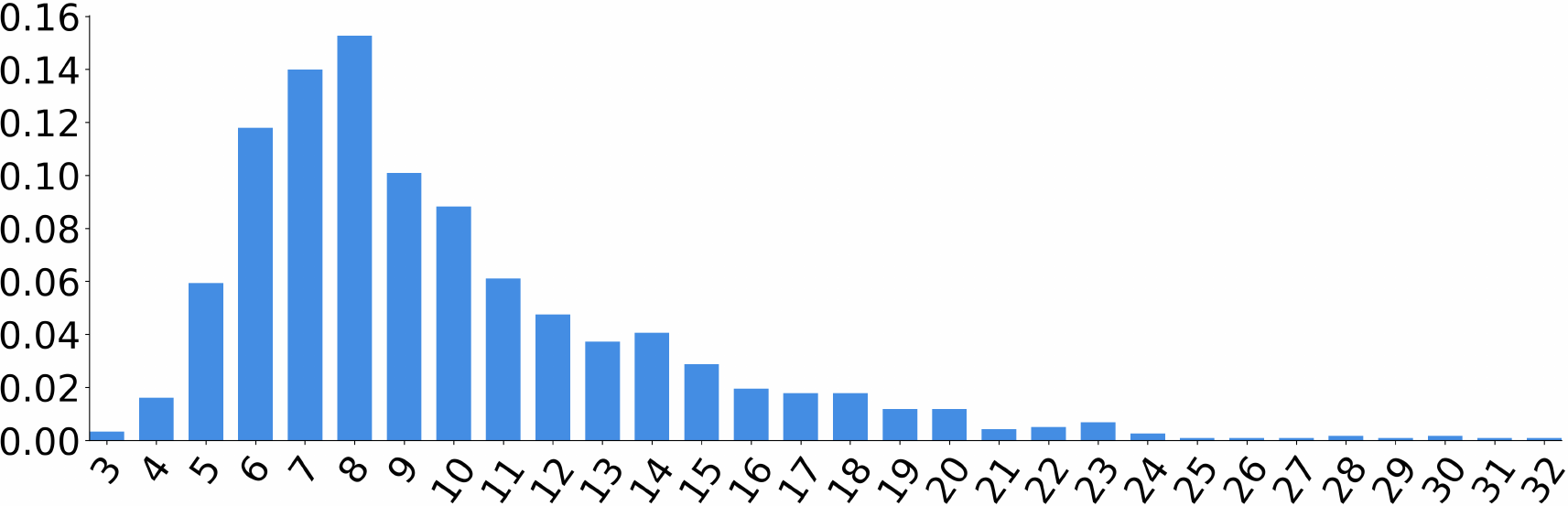}
    \caption{Distribution of task reference steps in WebRetriever, with x-axis as steps per trajectory and y-axis as task proportion (0–1 scale).}
    \label{figure:length}
\end{figure*}
We provide additional details on WebRetriever from multiple perspectives. Designed to reflect the heterogeneous nature of the global web, the dataset ensures evaluations capture diverse platforms and regions rather than being dominated by a few major markets. As shown in \cref{figure:info}(a), WebRetriever covers both cross-border platforms and region-specific services, including major ecosystems such as China and the US, as well as other geographic regions, including Europe (e.g., France, Spain), Asia, Africa, and Latin America. Complementing this geographic coverage, \cref{figure:info}(b) shows the language distribution. In addition to English and Chinese, the dataset incorporates regional languages such as Spanish and French, enabling rigorous assessment of agents’ cross-lingual comprehension and interaction capabilities while addressing the predominance of English content in prior benchmarks. This diverse geographic and linguistic composition makes WebRetriever a realistic and challenging benchmark for evaluating web agents in global, multilingual scenarios.
\par
\begin{table}[!t]
  \centering
  \caption{Examples of tasks from the three protocols in the WebRetriever benchmark. Fifteen tasks are randomly sampled, with 5 from each protocol.}
  \label{table:table8}
  \scriptsize
  \begin{tabularx}{\linewidth}{@{} l X @{}}
    \toprule
    \textbf{Website} & \textbf{Task Description} \\
    \midrule
    \rowcolor{blue!15}
    \multicolumn{2}{@{} c}{\textbf{Protocol I}} \\  % 分组标题，@{} 消除多余边距
    www.wjx.cn & Look up an introduction to informed consent. \\
    www.datajobs.com & Find a data science position around Boston. \\
    www.kaggle.com & Recommend a BigQuery-type dataset on Kaggle that has a Usability Rating of 10 and the highest number of downloads. \\
    www.aol.com & On the AOL website, search for “the best gifts for women” and browse the list of videos updated in the past week with a duration of 5–20 minutes. \\
    www.ci.nii.ac.jp & Using the advanced search functionality, find books whose publication period is between 2021 and 2025, with the data source set to “CiNii Books”, and whose full‑text fields contain the term “Artificial Intelligence”. \\
    \midrule
    \rowcolor{blue!15}
    \multicolumn{2}{@{} c}{\textbf{Protocol II}} \\
    www.district.in & List the Hindi 2D action movies currently showing in Ahmedabad. \\
    www.va.gov & Find out whether the U.S. Department of Veterans Affairs provides guide dogs for blind veterans. \\
    www.backstage.com & Find jobs within 500 km of London, England that are suitable for women aged 20 to 50. \\
    www.nih.gov & Find information on brain health charts within research institutes related to aging studies. \\
    www.komoot.com & Find a hiking trail within 100 km of New York City that takes under 3 hours to complete, is easy, has paved paths, an elevation gain below 200 meters, and has cafés along the route. \\
    \midrule
    \rowcolor{blue!15}
    \multicolumn{2}{@{} c}{\textbf{Protocol III}} \\
    www.cwur.org & On the 2024 GLOBAL 2000 LIST BY THE CENTER FOR WORLD UNIVERSITY RANKINGS, identify which universities in France are ranked in the top 0.2\%. \\
    www.cbre.com & On the CBRE Global Prime Office Rent Tracker for Q3 2025, identify which city in Europe had the lowest office rent level and is expected to see rent increases going forward. \\
    www.deloitte.com & Filter Deloitte articles published before 2025 with Sector: Food, Topic: Marketing, and Industry: Consumer Products, list the titles of the articles, and identify the first author of each article. \\
    www.pewresearch.org & One year after the start of Donald Trump’s second term, besides healthcare, goods, and housing costs, what economic issue were Americans somewhat most concerned about? \\
    www.catalog.mit.edu & Find the Massachusetts Institute of Technology (MIT) Course Catalog for the 2021–2022 academic year, locate the Computer Science course numbered ‘6.006’ (Introduction to Algorithms), and list the course numbers of the explicitly required direct prerequisites in the course description. If prerequisites are connected with “or” relationships, retain the full logical expression. \\
    \bottomrule
  \end{tabularx}
\end{table}
Extending the analysis to task complexity, \cref{figure:length} illustrates the long-tailed distribution of reference steps across WebRetriever tasks. While many tasks involve moderate interactions, a substantial portion requires extended, multi-step operation sequences, highlighting the benchmark’s coverage of both simple and complex real-world navigation scenarios. This variability enables comprehensive evaluation of agents’ capabilities, from basic information retrieval to long-horizon planning. To further demonstrate the diversity of task structures and difficulty levels captured by WebRetriever, examples from all three protocols are provided in \cref{table:table8}.

\section{Experiment Details}
\subsection{Base Model}
We evaluate a set of web agents on the WebRetriever benchmark, including SeeAct, Browser-Use, UI-TARS-1.5, Agent-E, and the Computer Use modes of Claude-4.5-Sonnet and Gemini-2.5-Pro. The backbone model for SeeAct, Browser-Use, and Agent-E is gpt-4o-2024-08-06. For the automatic evaluation method, the backbone models are gpt-4o-2024-08-06, o4-mini-2025-04-16, and Claude-4.5-Sonnet. GPT-4o is configured with a temperature of 0, o4-mini uses its default reasoning effort level of medium, and Claude-4.5-Sonnet operates with standard settings.

\subsection{Observation Scope}
Different web agents adopt different viewpoints when interacting with webpages in automated tasks. Specifically, SeeAct and Agent-E operate under a full-page view, capturing the entire page content, whereas Browser-Use, UI-TARS-1.5, and the Computer Use modes of Claude-4.5 and Gemini-2.5-Pro operate under a visible-area view, interacting only with the portion of the page currently displayed on the screen.

\subsection{Evaluation Independence}
To ensure evaluation integrity, NavEval and human reviewers independently assessed the same agent trajectories. The NavEval development team and the human reviewers were completely independent, with no personnel overlap. Moreover, the human reviewers had no access to NavEval's outputs during their assessment. This separation ensures that the human agreement rates reported in the main text reflect genuine alignment rather than information leakage between the two evaluation pipelines.

\subsection{Ablation Study on NavEval}
\begin{table}[!h]
    \centering
    \caption{Ablation study of NavEval performance with different LLMs. Avg AR denotes the average human agreement rate. All values are reported as percentages (\%).}
    \label{table:table9}
    \small
    \resizebox{0.7\textwidth}{!}{%
    \begin{tabular}{lccccc}
    \toprule
    \textbf{Method} & \textbf{Model} & \textbf{SeeAct} & \textbf{Agent-E} & \textbf{Browser-Use} & \textbf{Avg AR} \\
    \midrule
    \multirow{3}{*}{NavEval} 
     & GPT-4o & 87.6 & 86.1 & 85.6 & 86.4 \\
     & O4-mini & 89.9 & 88.7 & 88.5 & 89.0 \\
     & Claude-4.5-Sonnet & \cellcolor{blue!10}\textbf{92.2} &  \cellcolor{blue!10}\textbf{91.3} & \cellcolor{blue!10}\textbf{90.9} & \cellcolor{blue!10}\textbf{91.5} \\
    \bottomrule
    \end{tabular}
    }
\end{table}
In the main text, we noted that NavEval’s judgment module uses Claude-4.5-Sonnet to ensure stable semantic understanding and reasoning. To evaluate the impact of this design choice and assess the robustness and generalizability of the framework, we conduct an ablation study on NavEval’s LLM backbone. Specifically, while keeping all other components unchanged, we replace Claude-4.5-Sonnet with GPT-4o and O4-mini, and measure the resulting changes in task evaluation agreement with human judgments.
\cref{table:table9} summarizes the effect of different LLM backbones on NavEval. Claude-4.5-Sonnet achieves the highest agreement with human judgments, with an average AR of 91.5\%, while replacing it with O4-mini or GPT-4o results in modest drops to 89\% and 86.4\%, respectively. Agent-specific results follow the same trend. These findings indicate that NavEval is robust across LLM backbones, maintaining high performance overall, with Claude-4.5-Sonnet providing the best semantic reasoning and alignment with human evaluation.
\par
\begin{table}[!h]
    \centering
    \caption{Ablation study on rule-based filtering in NavEval. Avg AR denotes the average human agreement rate. All values are reported as percentages (\%). Settings indicate whether rule-based filtering is applied (w/ or w/o Filter).}
    \label{table:table10}
    \small
    \resizebox{0.95\textwidth}{!}{%
    \begin{tabular}{lccccccc}
    \toprule
    \textbf{Method} & \textbf{Setting} & \textbf{SeeAct} & \textbf{Agent-E} & \textbf{Browser-Use} &
    \textbf{\makecell{Gemini-2.5-Pro\\(Computer-Use)}} & \textbf{\makecell{Claude-4.5\\(Computer-Use)}} & \textbf{Avg AR} \\
    \midrule
    \multirow{2}{*}{NavEval} 
     & w/o Filter & 87.8 & 88.6 & 87.5 & 87.9 & 87.1 & 87.8 \\
     & w/ Filter & \cellcolor{blue!10}\textbf{92.2} & \cellcolor{blue!10}\textbf{91.3} & \cellcolor{blue!10}\textbf{90.9} & \cellcolor{blue!10}\textbf{91.4} & \cellcolor{blue!10}\textbf{90.1} & \cellcolor{blue!10}\textbf{91.2} \\
    \bottomrule
    \end{tabular}
    }
\end{table}
We further analyze the role of rule-based filtering within NavEval by removing this component while keeping all other settings unchanged. As shown in \cref{table:table10}, removing rule-based filtering reduces NavEval's Avg AR from 91.2\% to 87.8\%, a drop of 3.4 percentage points. Nevertheless, even without filtering, NavEval still outperforms the best prior method (WebJudge at 81.0\% in the main text), suggesting that the primary performance gains primarily come from the network-request signal itself rather than the filtering heuristics. The filtering module instead serves as a refinement step, reducing noise and normalizing payloads, which leads to consistent improvements across all evaluated agents.

\section{Case Study}
In this section, we present representative case studies from the WebRetriever benchmark, using the Computer Use mode of Gemini-2.5-Pro to illustrate agent performance across scenarios of varying complexity.

\subsection{Pop-up Window}
\begin{figure}[!b]
    \centering
    \includegraphics[width=0.5\linewidth]{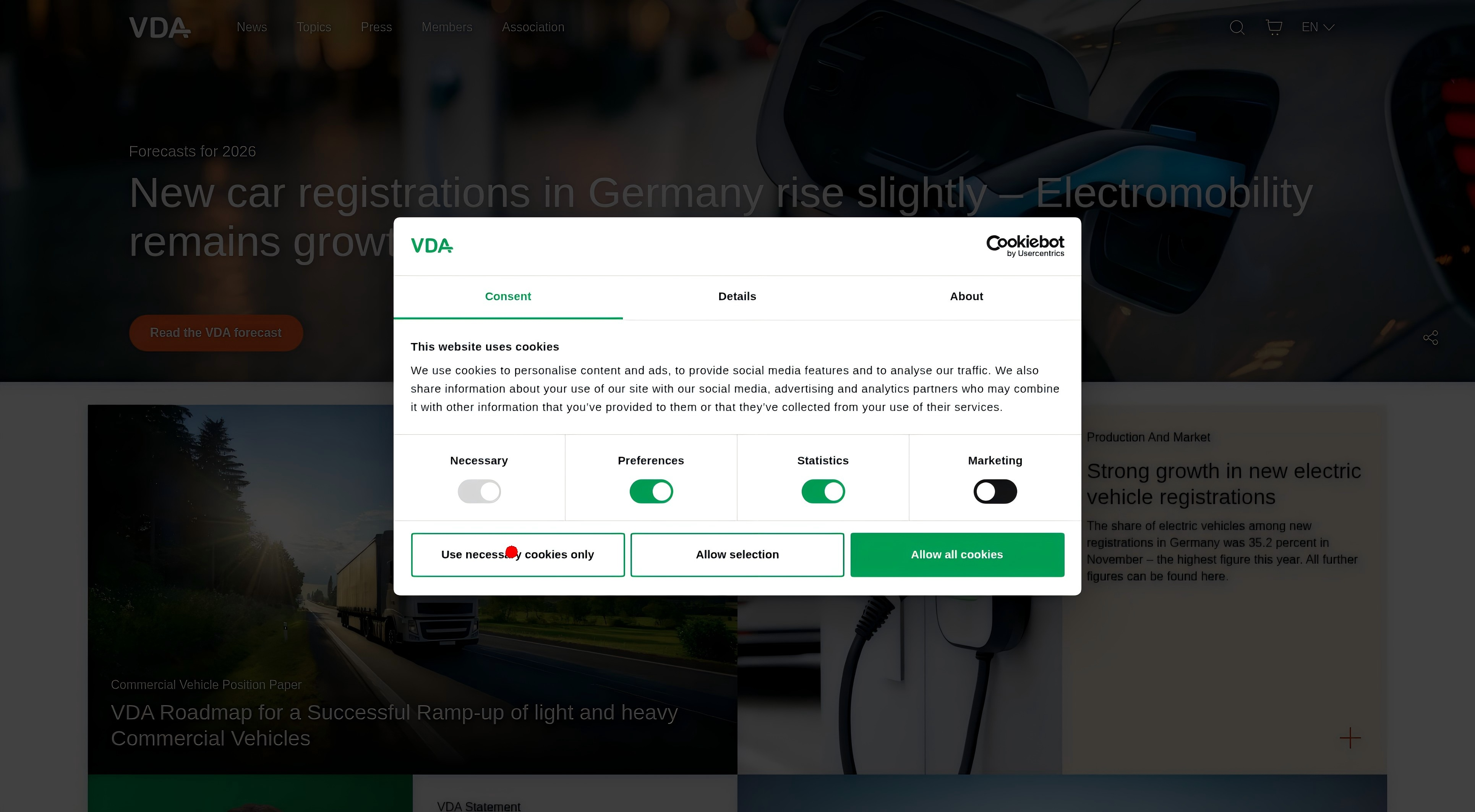}
    \caption{Task: "Find the introduction to autonomous driving functions provided by VDA."}
    \label{figure:case1}
\end{figure}

\cref{figure:case1} shows a task involving unexpected pop-up windows that appear during execution. Agents must interact with these elements to access hidden information.

\subsection{Collapsible Content}
\begin{figure}[!t]
    \centering
    \begin{subfigure}{0.48\linewidth}
        \centering
        \includegraphics[width=\linewidth]{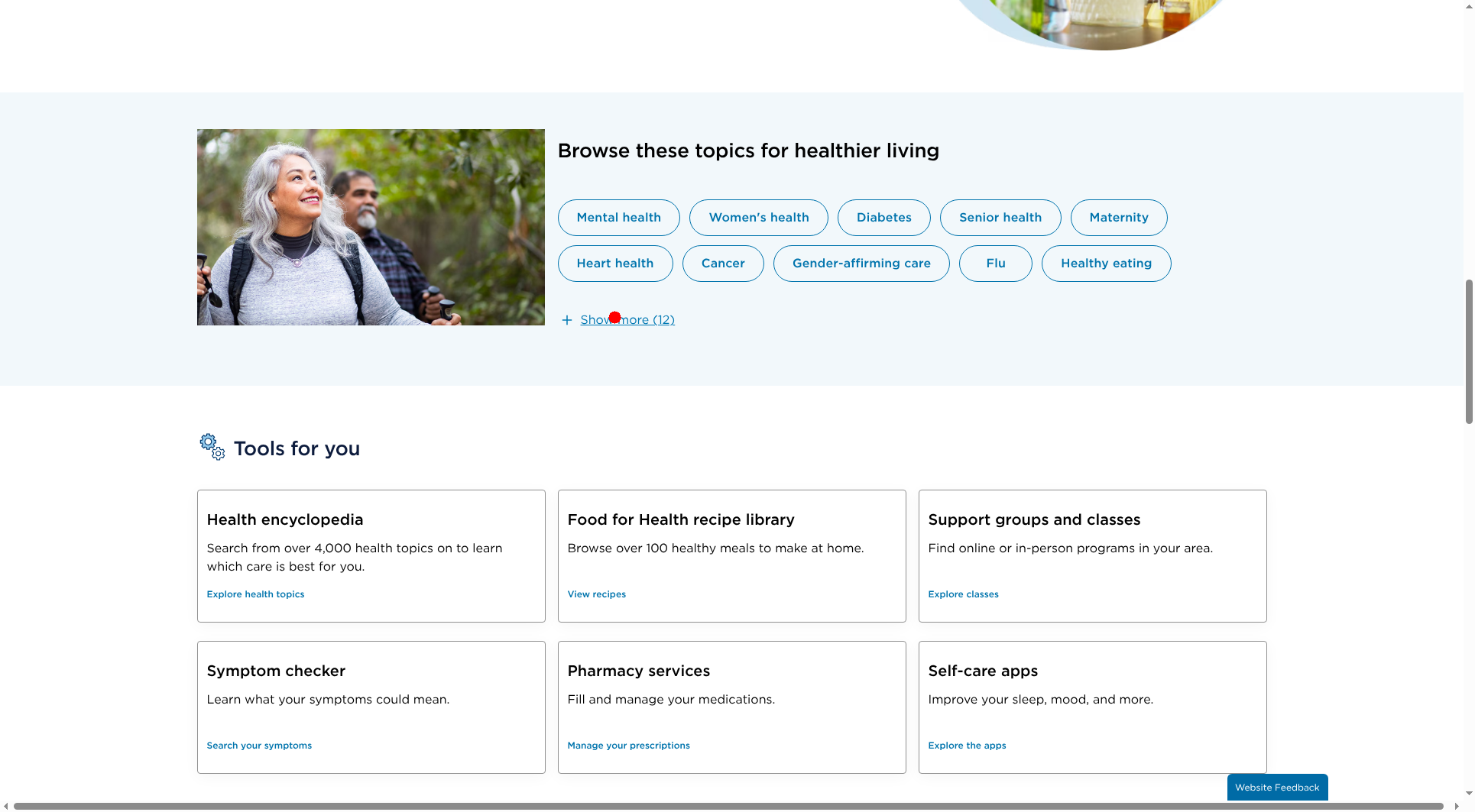}
        \caption{Click "Show more" to display additional information.}
    \end{subfigure}
    \begin{subfigure}{0.48\linewidth}
        \centering
        \includegraphics[width=\linewidth]{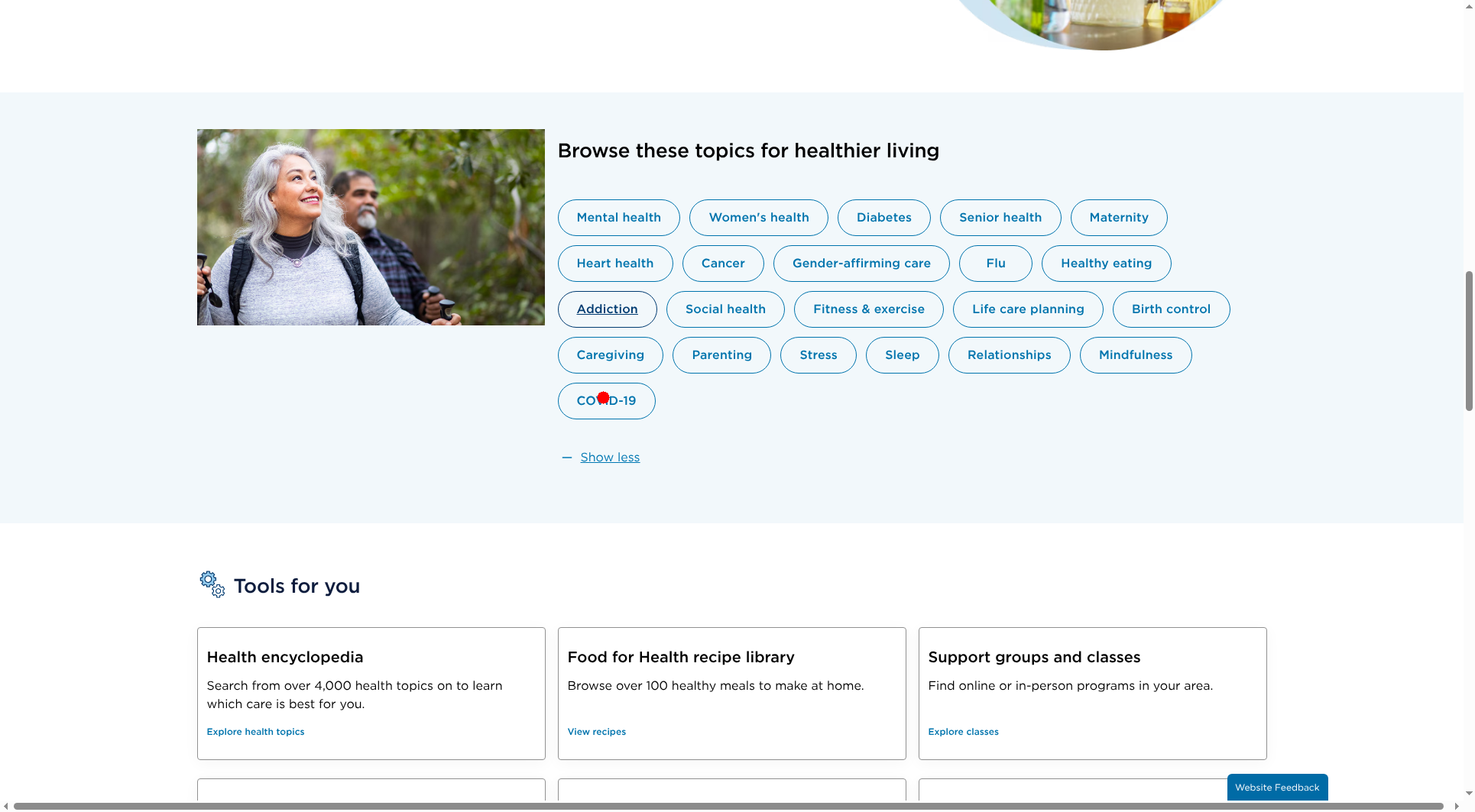}
        \caption{Click "COVID-19" to view available vaccination locations.}
    \end{subfigure}
    \caption{Task: "List the locations in Washington where vaccines are available".}
    \label{figure:case2}
\end{figure}

\cref{figure:case2} presents tasks with collapsible sections. Agents are required to expand hidden content to retrieve the relevant information.

\subsection{Alphabetical Index}
\begin{figure}[!t]
    \centering
    \begin{subfigure}{0.48\linewidth}
        \centering
        \includegraphics[width=\linewidth]{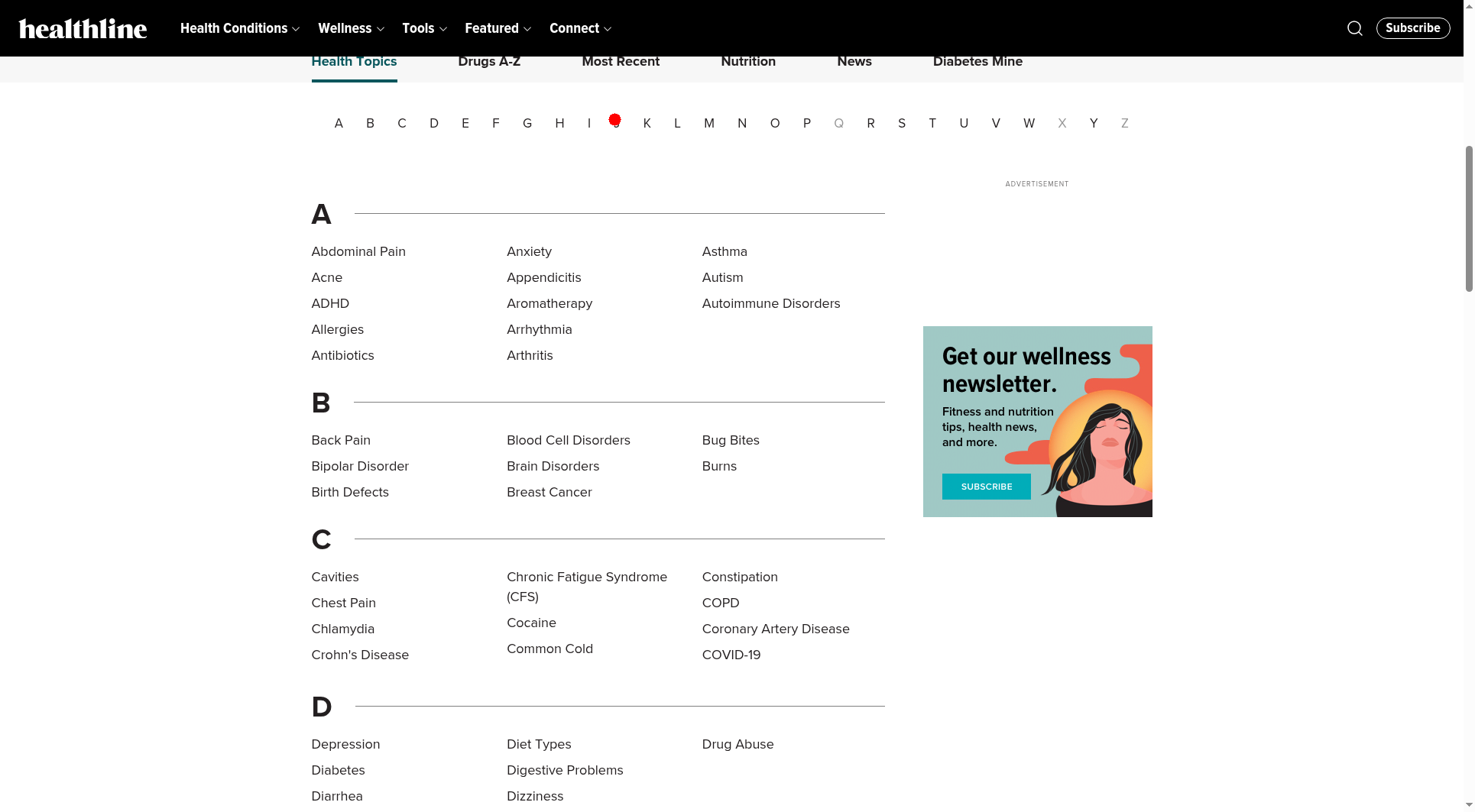}
        \caption{Click the 'J' category on the A–Z index page to further browse related entries.}
    \end{subfigure}
    \begin{subfigure}{0.48\linewidth}
        \centering
        \includegraphics[width=\linewidth]{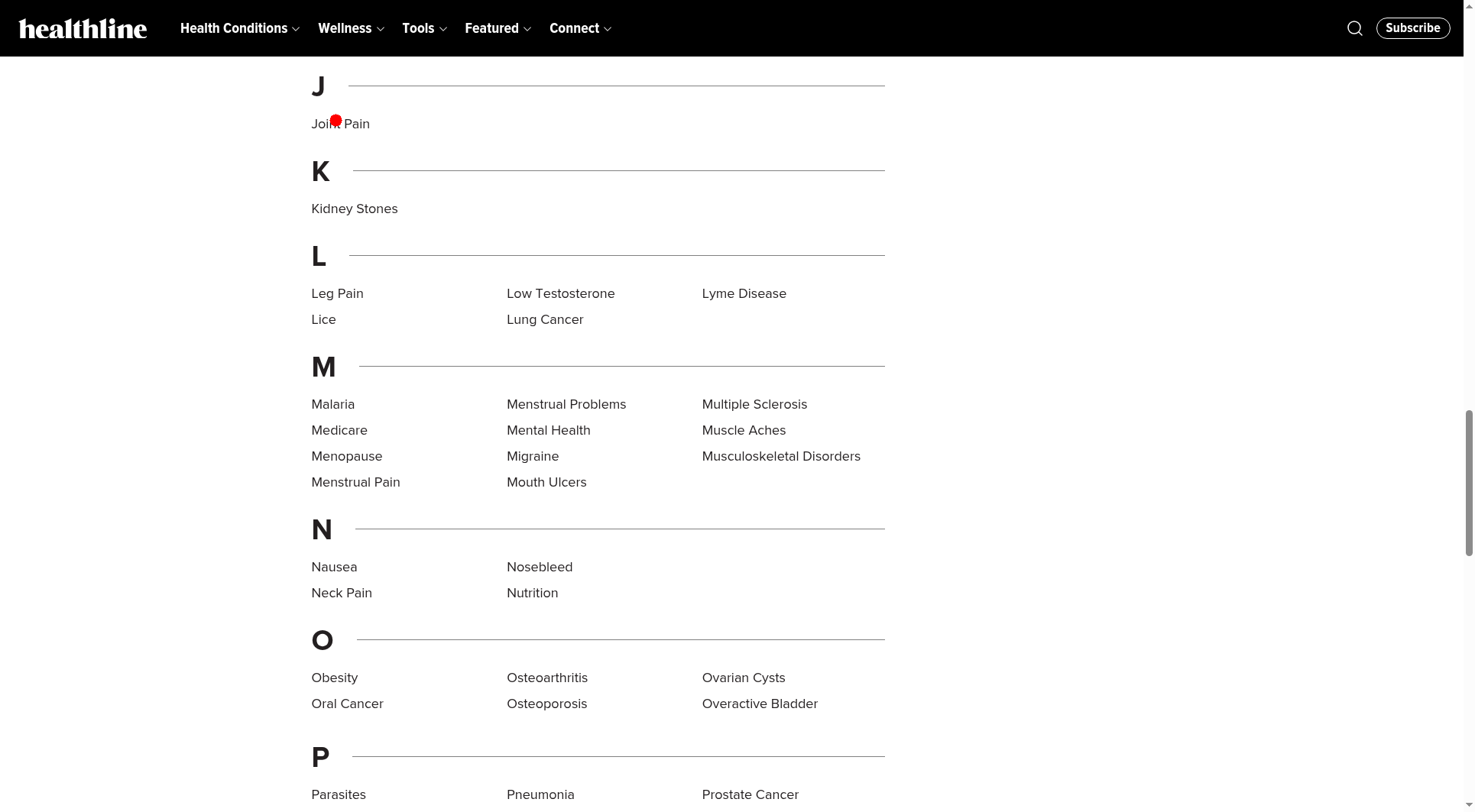}
        \caption{Click 'Joint Pain' on the 'J' category page to view the possible causes.}
    \end{subfigure}
    \caption{Task: "Find information about the possible causes of joint pain".}
    \label{figure:case3}
\end{figure}

\cref{figure:case3} illustrates tasks with information organized by an alphabetical index, requiring agents to navigate to specific entries and retrieve targeted information.

\subsection{Hallucinations}
\begin{figure}[!t]
    \centering
    \includegraphics[width=0.55\linewidth]{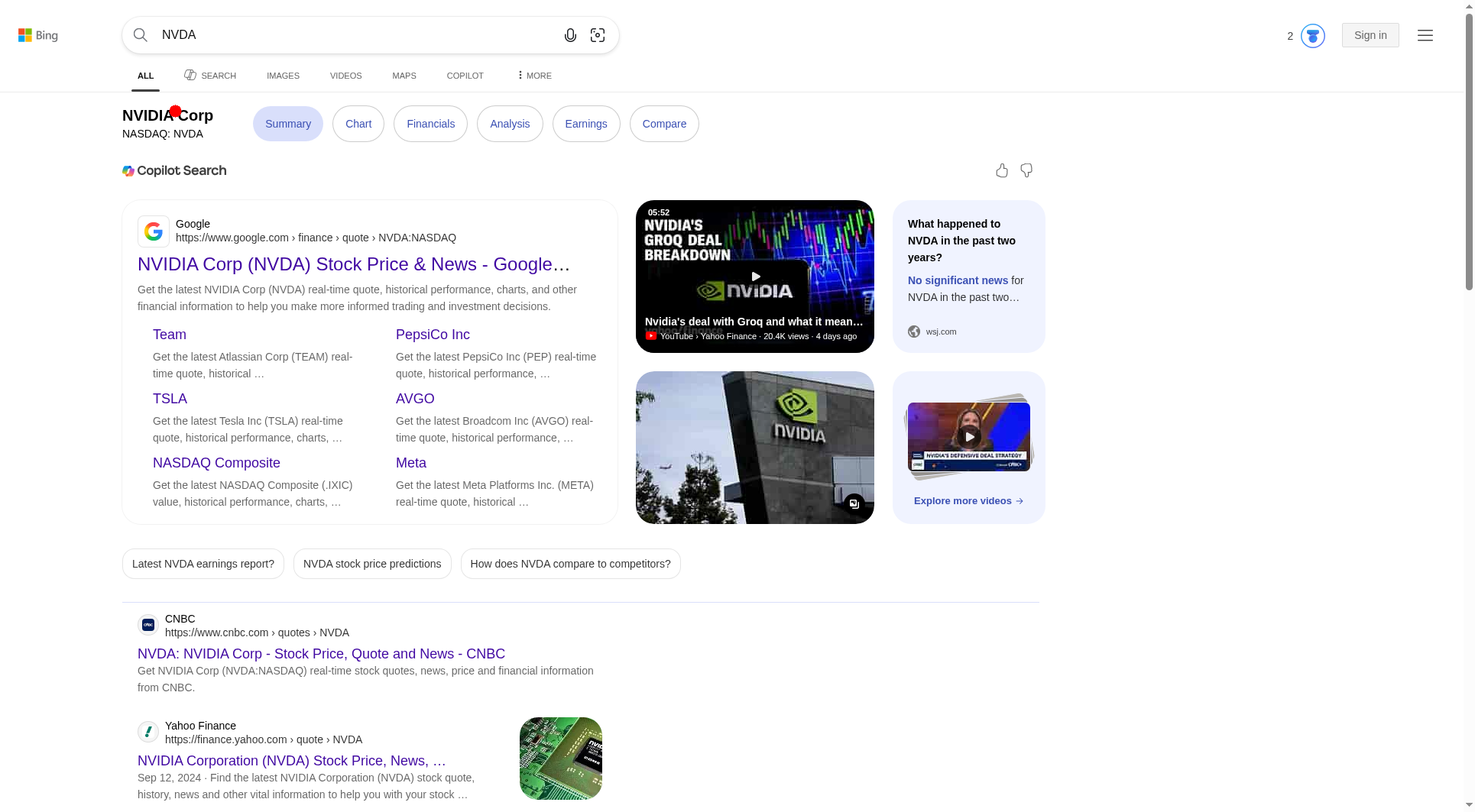}
    \caption{Task: "View a comparative trend chart showing the year-to-date stock performance of NVIDIA (NVDA) and Microsoft (MSFT)."}
    \label{figure:case4_1}
\end{figure}

The Web agent occasionally exhibits hallucinations, performing unnecessary or incorrect interactions. In \cref{figure:case4_1}, the reasoning process for the executed action is as follows:
\textit{"The current page already displays the relevant information and search results for NVIDIA Corp (NVDA), but in order to view more detailed company information and stock trends, it is necessary to enter its detailed stock page. Clicking “NVIDIA Corp” can enter the company’s detailed information page, thereby obtaining more information about its stock performance, analysis, and comparison options"}.
However, as shown in \cref{figure:case4_1}, the agent hallucinates and treats the click location as an interactive component.
\par
\begin{figure}[!t]
    \centering
    \includegraphics[width=0.55\linewidth]{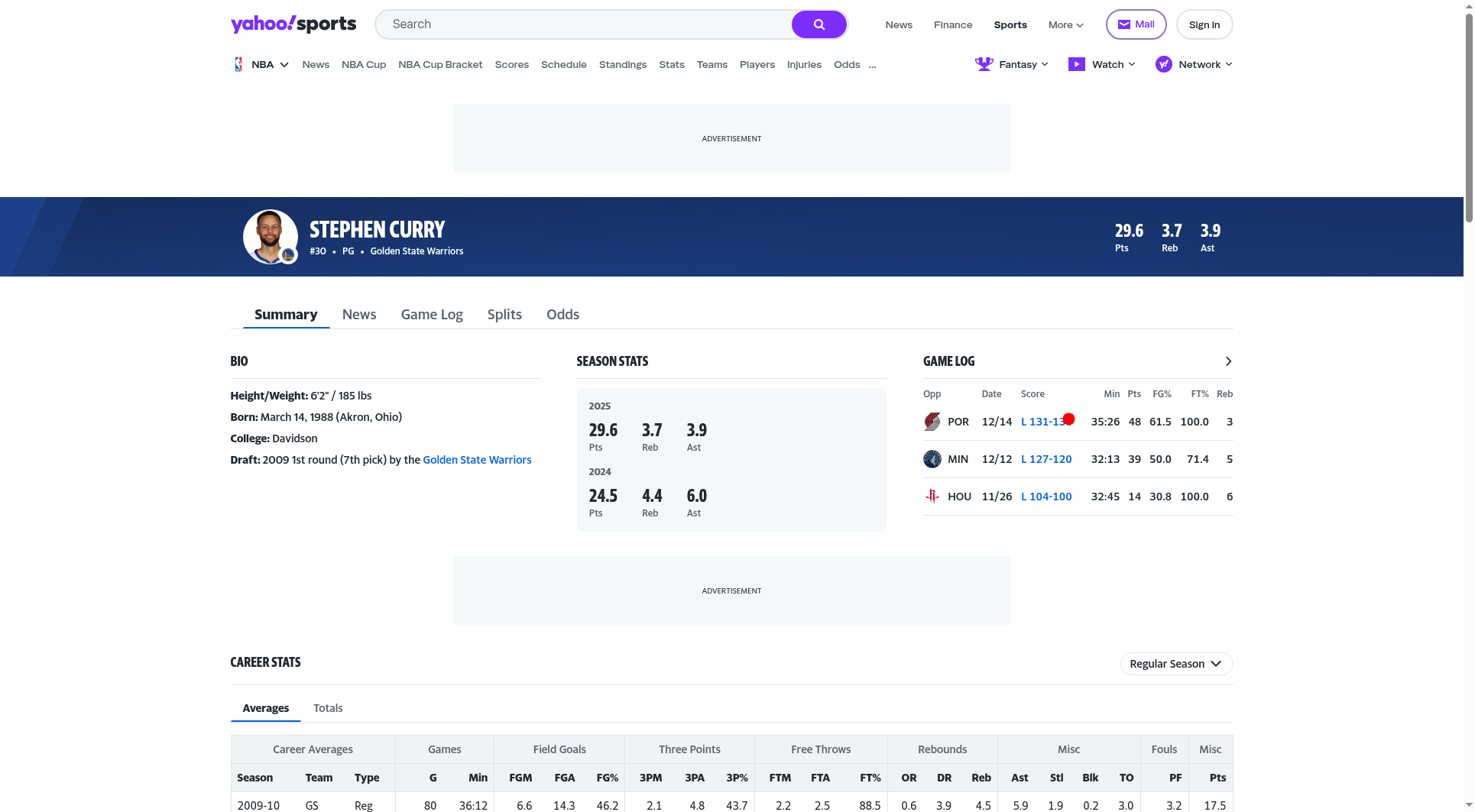}
    \caption{Task: "View the scoring performance of Stephen Curry in his most recent game."}
    \label{figure:case4_2}
\end{figure}

Similarly, in \cref{figure:case4_2}, the reasoning process is:
\textit{"The current page already displays detailed information about Stephen Curry, including basic information, season statistics, and the game log. In order to view the scoring situation of his most recent game, it is necessary to find the first game in the “Game Log” section and click the corresponding “Score” data. In this way, the detailed scoring information of that game can be obtained"}. 
Yet, as shown in \cref{figure:case4_2}, the score of the most recent game is already visible, and the agent still hallucinates by attempting an unnecessary interaction.

\section{Prompt Template}
\label{sec:prompt}

\newtcolorbox{promptbox}[1]{
  enhanced,
  breakable,                % 允许跨页
  title={#1},               % 标题
  % --- 颜色调整 (小清新风格) ---
  colframe=teal!65!white,   % 边框：柔和的蓝绿色
  colback=teal!4!white,     % 背景：极淡的薄荷白
  % ---------------------------
  coltitle=white,           % 标题文字颜色
  fonttitle=\bfseries\normalsize,% 标题字体
  fontupper=\small,
  % halign title=center,
  boxrule=1pt,              % 边框线条宽度
  toptitle=1mm,             % <--- 关键：标题文字上方留空
  % bottomtitle=1mm,          % <--- 关键：标题文字下方留空
}

\subsection{NavEval Evaluation}
 \label{sec:prompt:naveval}
 
\begin{promptbox}{System Prompt:}
    \par
    You are a benchmark validator.  
    Your job is to decide whether a browser-automation agent successfully completed a task by examining: \\
    
    \par\medskip
    
    1. Task description – defines the goal and the key actions required. Every requirement, keyword, constraint, or objective in the task is treated as a criterion.  \\
    2. Network requests – simplified logs revealing user operations such as searches, filters, sorts, form submissions, item selections, report retrieval, etc.  \\
    3. Final screenshot – shows the final state of the page and is used to verify that the result matches the task requirements.  \\
    4. URL trajectory – the full ordered list of URLs visited during navigation, reflecting the steps taken to reach the final state.  \\

    \par\medskip
    
    You must use the combination of these three evidence channels to evaluate whether each key requirement in the task was fulfilled.  
    However, if a single evidence source alone is already sufficient to confirm or reject a requirement, you may rely on that source. \\
    
    \par\medskip
    \par\medskip
    
    ** Evaluation Principles \\
    
    *** 1. Break the task into atomic requirements  \\
    Examples include: filters, ranges, keywords, sort criteria, extremum criteria (cheapest, highest-rated, newest, etc.), selecting a specific item, opening a report, completing a submission, or reaching a particular informational view.
    Each requirement becomes one action item in the output. \\

    \par\medskip
    \par\medskip
    
    *** 2. Evidence interpretation \\
    
    **** Network requests  
    Requests reveal operational steps such as:  \\
    - applied filters  \\
    - applied sort orders  \\
    - search queries  \\
    - backend apis  \\
    - item detail requests (ASINs, product IDs, listing IDs)  \\
    - “submit”, “search”, “apply”, or other key transitions  \\
    
    Requests are often the clearest indicator of whether a requirement was explicitly performed.  \\

    \par\medskip
    
    **** URL trajectory  
    URL history is used to determine the sequence of operations, including:
    
    - when filtering occurred  \\
    - when an item was opened  \\
    - when a page transition corresponds to a specific step in the task  \\
    - whether the sequence of URLs aligns with the required workflow  \\

    In some tasks, URL sequence contains the expected operational milestones implied by the task.  \\

    \par\medskip
    
    **** Final screenshot  
    Used to verify the final state:
    
    - correct result set displayed  \\
    - correct item or report visible  \\
    - correct filter values reflected visually  \\
    - correct summary or confirmation shown  \\
    - correct price, date, rating, or other constraints satisfied  \\
    
    Screenshot = state verification; requests/URLs = operation verification.  \\

    \par\medskip
    \par\medskip
    
    *** 3. Strict rules for numeric / filter / sort requirements \\
    
    If the task defines a numeric range or exact condition (price, year, beds, dates, etc.), the requirement must match exactly:
    
    - `\$1500–\$2500` requires that exact interval  \\
    - “exactly 2 beds” cannot be satisfied by “2+ beds”  \\
    - “newest”, “cheapest”, “highest-rated”, etc. require a sort or otherwise unambiguous extremum selection  \\
    
    Failure to match these conditions exactly → requirement fails.  \\

    \par\medskip
    \par\medskip
    
    *** 4. Success, failure, confidence \\
    You must output:
    - Whether all requirements were satisfied  \\
    - A confidence score  \\
    - Per-requirement judgments  \\
    
    Success (all\_passed = true)  \\
    All key requirements are supported by evidence. \\
    
    Failure (all\_passed = false)  \\
    At least one requirement is violated or cannot be confidently verified. \\
    
    Confidence scoring:
    
    - Clear success: 0.75–1.0  \\
    - Clear failure: 0.10–0.39  \\
    - Evidence insufficient to confirm a requirement: 
      failure with 0.40–0.50, and explanation why evidence does not conclusively satisfy the requirement. \\

    \par\medskip
    \par\medskip

    ** Output Format (Strict)

    \begin{lstlisting}[breaklines=true, basicstyle=\ttfamily\small]
    {
        "all_passed": True/False,
        "score": <float number between 0-1>,
        "reasoning": <top level reasoning why the result is success or failure>,
        "details": [
            {
                "action": <key action to validate>,
                "passed": True/False,
                "reason": <detailed reason>
            }
        ]
    }
    \end{lstlisting}

    \par\medskip

    Return only the JSON object, no commentary.
    
\end{promptbox}

\begin{promptbox}{User Prompt:}
    \par
    Task: <task> \\
    Url Trajectory: <url\_trajectory> \\
    Network Requests: <network\_requests> \\
    Last Screenshot: <last\_image>
\end{promptbox}

\subsection{Generation Operational Documentation}
\label{sec:prompt:documention}

\begin{promptbox}{System Prompt:}
    \par
    You are an expert at converting web automation trajectories into ONE explicit English key path sentence.  \\

    \par\medskip
    
    You are given a task title (the goal) and a sequence of UI actions. 
    Your output MUST be aligned with the task goal: the key path must describe how to accomplish the goal stated in the title.  \\

    \par\medskip
    
    ** Hard rules (must follow):  \\
    1. Output ONLY one English sentence. No quotes, no bullets, no numbering.  \\
    2. The sentence MUST follow the task goal (title) as the primary constraint. Do NOT mechanically list actions; instead, describe actions as steps that move toward completing the goal.  \\
    3. You must reflect ALL meaningful action types that appear in the trajectory, including Click/Type/Select/Toggle/ScrollDown/ScrollUp.  \\
    4. Scroll actions MUST NOT be omitted and MUST be written with a goal/purpose inferred from the task goal and the subsequent non-scroll action.  \\
    5. Preserve the real execution order of actions, but compress redundant micro-steps when possible.  \\
    
    \par\medskip
    
    ** Click/selection wording constraints:  \\
    1. When describing what to click/select, prefer describing the UI location/position/role (e.g., navigation entry, search result position, list position, tab, filter option) rather than copying long content text.  \\
    2. Avoid quoting full article/news/product titles or long sentences from the page. If a label is necessary, keep it short and functional.  \\
    3. If multiple candidates exist on the page, express which one by ordinal/position (first/second/top) or by function, so the instruction is goal-directed and reproducible.  \\
    
    \par\medskip
    
    ** Output constraints:  \\
    1. Do NOT include coordinates, bbox, ids, file paths, or step numbers.  \\
    2. Do NOT invent actions that do not exist in the trajectory.  \\
    3. Prefer a structured flow such as “first… then… next… during… finally…”, but keep it ONE sentence.  \\
    4. Keep under 100 English characters if possible; correctness and goal-alignment are more important than brevity.  \\
    5. Preserve important UI labels/options in original casing if they are in English. 
    
\end{promptbox}

\begin{promptbox}{User Prompt:}
    \par
    Task: <task> \\
    Website: <website> \\
    Trajectory: <trajectory>
    \par
    \text{}
    \par
    Now generate the operational document sentence:
\end{promptbox}

\end{document}